\documentclass{article}
\PassOptionsToPackage{numbers, compress}{natbib}

\usepackage[preprint]{neurips_data_2023}

\usepackage[utf8]{inputenc} 
\usepackage[T1]{fontenc}    
\usepackage{hyperref}       
\usepackage{url}            
\usepackage{booktabs}       
\usepackage{amsfonts}       
\usepackage{nicefrac}       
\usepackage{microtype}      
\usepackage{xcolor}         
\usepackage{longtable}
\usepackage{graphicx}
\usepackage{array}
\usepackage{multirow}
\usepackage{xcolor,colortbl}
\usepackage{xspace}
\usepackage{makecell}
\usepackage{boldline}
\usepackage{pifont}
\usepackage{tabularx}
\usepackage{graphicx}
\usepackage{wrapfig}
\usepackage[export]{adjustbox}
\usepackage{fancyvrb}
\usepackage{fvextra}
\usepackage[most,breakable]{tcolorbox}
\usepackage{caption}
\usepackage{pgfplots}
\pgfplotsset{compat=newest}
\usepgfplotslibrary{units}
\usepackage{placeins}
\usepackage{lineno}

\newcommand{\ours}[0]{\textsc{AgentGen}\xspace}
\newcommand{\goal}[0]{\textsc{Bi-Evol}\xspace}
\newcommand{\framework}[0]{\textsc{AgentGen}\xspace}

\title{
\ours: Enhancing Planning Abilities for Large Language Model based Agent via Environment and Task Generation
}

\author{\textbf{Mengkang Hu}$^1$,
        \textbf{Pu Zhao}$^2$,
        \textbf{Can Xu}$^2$,
        \textbf{Qingfeng Sun}$^2$,
        \textbf{Jianguang Lou}$^2$,
        \textbf{Qingwei Lin}$^2$,\\ 
        \textbf{Ping Luo}$^1$,
        \textbf{Saravan Rajmohan}$^2$ \\
  $^1$The University of Hong Kong 
  \quad
  $^2$ Microsoft Corporation \\
  \texttt{\{v-humengkang,puzhao\}@microsoft.com},~~\texttt{pluo.lhi@gmail.com},\\
  \texttt{\{caxu,qins,jlou,qlin,saravar\}@microsoft.com}
  \vspace{-0.5cm}
}

\begin{document}

\maketitle

\begin{abstract} 

Large Language Model (LLM) based agents have garnered significant attention and are becoming increasingly popular. 
Furthermore, \textit{planning} ability is a crucial component of an LLM-based agent, involving interaction with the \textit{environment} and executing actions to complete a \textit{planning task}, which generally entails achieving a desired goal from an initial state.
This paper investigates enhancing the planning abilities of LLM-based agents through instruction tuning, referred to as \textit{agent training}.
Recent studies on agent training have demonstrated that utilizing expert-level trajectory data (sequences of action-observation pairs) for instruction-tuning LLMs effectively enhances their planning capabilities.
However, existing work primarily focuses on synthesizing trajectories from manually designed planning tasks and environments. The labor-intensive nature of creating these environments and tasks impedes the generation of sufficiently varied and extensive trajectories for agent training.
To address this limitation, this paper explores the automated synthesis of diverse environments and a gradual range of planning tasks, from easy to difficult. 
We introduce a framework, \ours, that leverages LLMs first to generate environments and subsequently generate planning tasks conditioned on these environments.
Specifically, to improve \textit{environmental diversity}, we propose using an inspiration corpus composed of various domain-specific text segments as the context for synthesizing environments. 
Moreover, to increase the \textit{difficulty diversity} of generated planning tasks, we propose a bidirectional evolution method, \goal, that evolves planning tasks from easier and harder directions to synthesize a task set with a smoother difficulty curve, thereby enhancing the learning process of LLMs more effectively.
These methods collectively contribute to the generation of diverse trajectory data for instruction-tuning.
Based on \framework, we greatly expanded the number of environments and planning tasks available for agent training.
The evaluation results from AgentBoard indicate that \framework greatly enhances the planning capabilities of LLMs. 
For instance, the \ours instruction-tuned Llama-3.1-8B outperforms GPT-3.5 in overall performance. 
Moreover, the \ours-tuned Llama-3.1-70B model achieves state-of-the-art results in planning tasks.
Project page: \href{https://agent-gen.github.io/}{this URL}.
\end{abstract}

\section{Introduction}
\label{sec:intro}

Recently, owing to advancements in Large Language Models (LLMs)~\cite{opeiai2022gpt,openai2023gpt4,meta2024llama3,touvron2023llama}, the LLM-based Agents have garnered widespread attention from the artificial intelligence community. 
Generally, an LLM-based agent refers to utilizing LLMs to perceive the environment, make decisions, and execute actions to substitute or help people accomplish some specific tasks~\cite{xi2023rise,wang2023survey,TianbaoXie2023_OpenAgents}.
Furthermore, \textit{planning} is often regarded as one of the most important applications of LLM-based agents, such as robotic planning~\cite{shridhar2020alfworld,puig2018virtualhome,huang2022language,valmeekam2024planbench}, travel planning~\cite{zheng2024natural,xie2024travelplanner}, etc. 
In this study, planning is conceptualized as the systematic process of identifying a sequence of executable actions within a given \textit{environment} to complete a \textit{planning task}, defined as the transition from an initial state to achieve specified goal conditions, considering constraints and available resources~\cite{kaelblingTAMP,russell2016artificial}.

\begin{figure}[htbp]
    \centering
    \includegraphics[width=\linewidth]{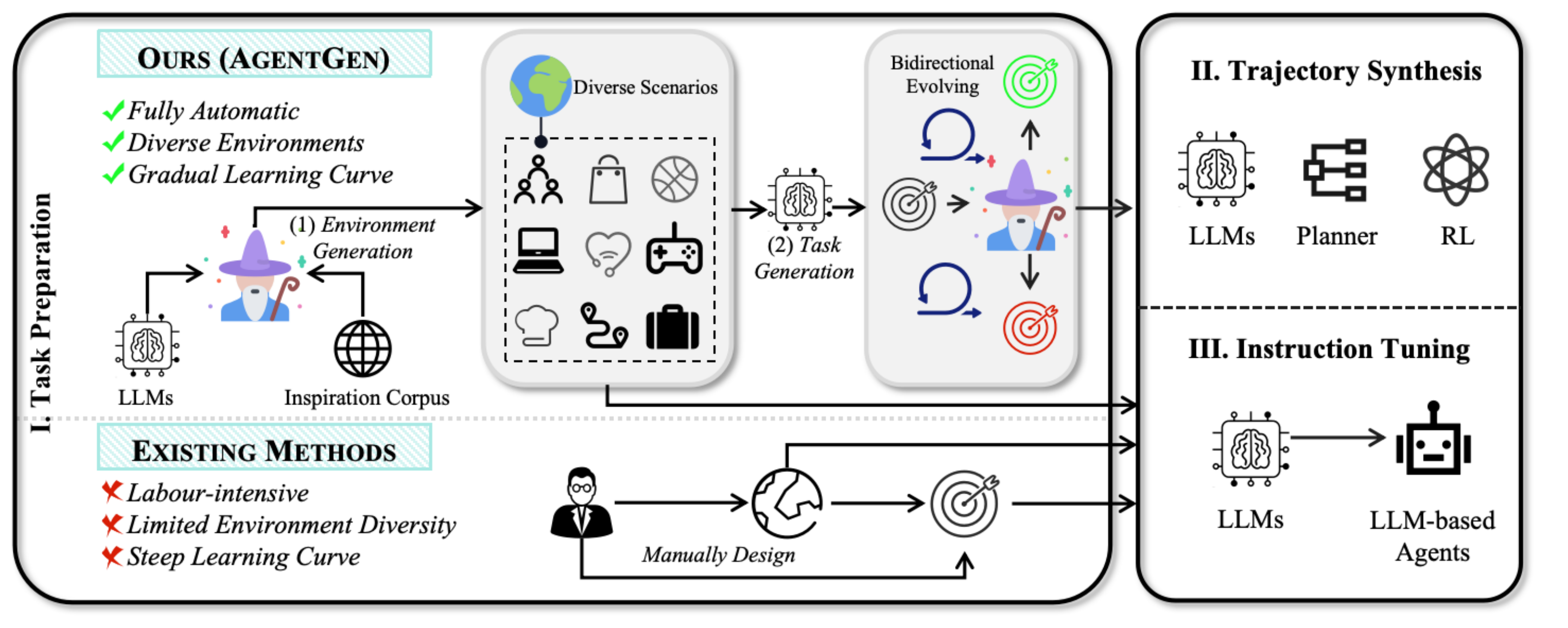}
    \caption{
    A typical agent training process includes three stages: task preparation, trajectory synthesis, and instruction tuning. \ours primarily distinguishes itself from existing agent training literature in the task preparation stage, where we introduce a \textit{fully automated} task generation framework \framework for constructing \textit{diverse} environments and planning tasks with \textit{gradual learning curves}.
    }
    \label{fig:teaser}
\end{figure}

Improving planning capabilities through instruction-tuning LLMs is a significant research problem, referred to as \textit{agent training}.
As shown in Figure~\ref{fig:teaser}, similar to imitation learning~\cite{hussein2017imitation}, a typical agent training process can be divided into three stages:
\textit{(i)} Preparing environments and planning tasks.
\textit{(ii)} Synthesizing expert-level trajectories (sequences of action-observation pairs) on these planning tasks. For example, utilizing state-of-the-art LLMs (e.g., GPT-4~\cite{openai2023gpt4}) as the agent and filtering trajectory based on reward score~\cite{zeng2023agenttuning,chen2023fireact}.
\textit{(iii)} Instruction-tuning LLMs with the synthesized trajectory data.
Recently, the effectiveness of enhancing the planning capabilities of LLMs through agent training has been demonstrated by many studies~\cite{zeng2023agenttuning,yin2023lumos,chen2023fireact,wang2024learning,chen2024agent,zhang2024agentohana,wang2024llms,song2024trial}.
Despite their success, one key limitation of these works is that they primarily rely on manually designed environments and planning tasks. 
The labor-intensive nature of creating environments and planning tasks hinders the generation of diverse and extensive trajectory data.
More explicitly, designing diverse environments requires defining a range of rich and practical scenarios, and implementing these environments typically involves the participation of human experts with programming skills.
Additionally, formulating tasks often demands creating a task set with a gradual difficulty progression.
Due to this constraint, existing agent training studies typically use only a few environments for data synthesis.

To address the aforementioned deficiencies, this paper introduces an automatic framework \textbf{\framework} that utilizes LLMs to construct diverse environments and planning tasks for agent training, expanding the available environments from a few to hundreds.
More specifically, \ours is structured around two stages:
(1) \textbf{\textit{Environment Generation}}: 
Achieving sufficient \textit{environmental diversity} is essential for creating diverse planning tasks, which involves covering a broad range of scenarios and domains. 
To ensure this, we use an \textit{inspiration corpus} composed of diverse text segments as context for generating environment specifications with LLMs, where actions, restrictions, and other details are defined using natural language.
For example, in Figure~\ref{fig:environment_generation}, we randomly selected a text segment from the inspiration corpus: \textit{``How to boost your diet with peanut butter powder?''} This prompted the generation of a related environment specification: \textit{``You are a nutritionist tasked with creating a new healthy recipe book that incorporates peanut butter powder as a key ingredient''}.
Subsequently, we prompt the LLM to produce the corresponding code based on this specification, which may be composed of Python, Planning Domain Definition Language (PDDL)~\cite{McDermott1998PDDLthePD}, or other domain-specific languages.
Furthermore, we constructed an environment library to serve as in-context examples and iteratively expanded it by incorporating high-quality newly generated environments.
(2) \textbf{\textit{Task Generation}}: Conditioned on the generated environment, we aim to create multiple planning tasks. 
In this stage, it is crucial to have a gradual set of tasks ranging from easy to difficult, i.e., \textit{difficulty diversity}.
To achieve greater difficulty diversity, we propose a bidirectional evolution method, \textbf{\goal}, where the LLM first generates random planning tasks and then evolves these tasks by applying constraints towards both simplification and increased difficulty. 
The created task set with \goal has a smooth difficulty curve that facilitates LLMs' smoother acquisition of planning skills.
To verify the effectiveness of our method, we synthesized environments and planning tasks based on PDDL~\cite{McDermott1998PDDLthePD} and constructed a dataset comprising 592 environments, each with 20 tasks. 
We then used a domain-independent planner to obtain 7,246 high-quality trajectories.
Subsequently, we used this trajectory data for instruction-tuning a series of LLMs and demonstrated the trained model on AgentBoard~\cite{ma2024agentboard}. 
Since our instruction-tuning dataset is composed of trajectory synthesized from PDDL-based planning tasks, we refer to evaluation tasks implemented in PDDL as \textit{in-domain tasks} and tasks implemented in other programming languages as \textit{out-of-domain tasks}. 
Importantly, this evaluation was conducted in a \textit{zero-shot} manner without utilizing any trajectory data from these tasks. 
Experimental results demonstrate that \ours achieved more than a tenfold improvement over the raw LLama-3.1-8B on in-domain tasks (33.3 vs. 3.0), with overall performance surpassing that of GPT-3.5. Furthermore, the performance of \ours-tuned Llama-3.1-70B exceeded GPT-4, setting a new state-of-the-art in planning tasks.
In out-of-domain tasks, \ours also demonstrated similar experimental outcomes. 
Specifically, it led to a significant improvement in average success rates, with the raw LLama-3.1-8B model achieving a 10.0\% increase and the 70B model a 3.7\% improvement.
In summary, the proposed environment and planning task generation method \ours can help improve planning ability. Moreover, not only can in-domain tasks benefit from this, but out-of-domain tasks also improve, which confirms both the effectiveness and generalization.
Our contributions can be summarized as follows:
\begin{itemize}
    \item We introduce \framework, which, as far as we know, is the first framework for automatically generating diverse planning tasks and environments targeted for LLM-based agent training.
    \item We propose utilizing an inspiration corpus as the context for generating environments with LLMs, resulting in 592 diverse environments that encompass a broad range of scenarios.
    \item We propose a bidirectional evolution method \goal that evolves seed planning tasks in both simpler and more complex directions, thereby constructing a task set with a smoother difficulty curve.
    \item We constructed an agent instruction-tuning dataset with 7246 high-quality trajectories through \framework. LLMs instruction-tuned with this dataset achieved massive improvement in both in-domain and out-of-domain planning tasks, which validated the effectiveness and generalization of \framework.

\end{itemize}

\section{Preliminary}
\label{sec:preliminary}
\vspace{-5pt}

\subsection{Planning Problems}
\label{sec:agent_task_definition}
We consider goal-directed deterministic planning problems~\cite{russell2016artificial}, which are formally defined as a tuple $\mathcal{P} = (\mathbb{T},~\mathbb{E})$, where $\mathbb{E}$ denotes the environment in which the agent interacts and $\mathbb{T}$ denotes the task that the agent needs to complete.
Specifically, an environment $\mathbb{E}$ typically models a world, encompassing the definitions of the action space $\mathcal{A}$ and state space $\mathcal{S}$, as well as the transition function $\mathcal{T}: \mathcal{S} \times \mathcal{A} \to \mathcal{S}$.
Task $\mathbb{T}$ is further defined by the tuple $\mathbb{T} = (\mathcal{G},~\mathcal{I})$, where $\mathcal{G}$ refers to the goal conditions and $\mathcal{I}$ refers to initial states of the agent.
The initial states $\mathcal{I}$ are a subset of the state space $\mathcal{S}_{i}$ that specifies the starting conditions of the agent.
The goal $\mathcal{G}$ is a subset of the state space $\mathcal{S}_{g}$ that specifies the desired outcomes or conditions. Specifically, $\mathcal{G}$ can be expressed as $\mathcal{G} = \{ s \in \mathcal{S}_{g} \mid \phi(s) = \text{true} \}$. Here, $\phi(s)$ is a boolean-valued function representing conditions or propositions that must be satisfied for the state $s$ to be considered part of the goal set. 

\subsection{Planning Problem Implementation}



A planning problem can be implemented with programming languages such as Python or domain-specific languages such as Planning Domain Definition Language (PDDL)~\cite{McDermott1998PDDLthePD}.
For example, in a PDDL-based planning problem, the domain PDDL file can be regarded as the environment $\mathbb{E}$, defining states (predicates) and actions and specifying the transition function using preconditions and effects of each action. The problem PDDL file, on the other hand, can be seen as the task $\mathbb{T}$. Both initial states and goal conditions are typically defined as combinations of predicates.
Another widely used programming language for constructing planning problems is Python.
For example, in OpenAI gym\footnote{\url{https://www.gymlibrary.dev/index.html}}, a planning problem will be implemented as a Python class, where the transition function is implemented as a method of the class, usually named the "step" or "update" function. Meanwhile, the goal $\mathcal{G}$ is typically represented as a reward function that indicates the objective of the task, and the initial states $\mathcal{I}$ are defined in a method named "reset."

\subsection{Large Language Model based Agent}
An LLM-based agent leverages a pre-trained language model to operate within the defined environment $\mathbb{E}$ and complete the given task $\mathbb{T}$.
Given an environment $\mathbb{E}$, the LLM-based agent perceives its state $\mathcal{S}$ and takes actions $\mathcal{A}$ based on its understanding and processing of the input. 
The transition function $\mathcal{T}: \mathcal{S} \times \mathcal{A} \to \mathcal{S}$ remains consistent, where the LLM-based agent determines the next state by generating appropriate actions through natural language processing.
The goal $G$ guides the LLM-based agent in selecting actions that maximize the reward. The agent utilizes the language model to interpret the task requirements and generate actions that align with achieving the specified goal.
In essence, the LLM-based agent forms a policy $\pi: \mathcal{S} \to \mathcal{A}$ using the LLM, where $\pi(s)$ is the action taken in state $s$ based on the LLM's understanding and processing of the task.

\section{Methodology}

\paragraph{Problem Definition}
\label{sec:problem_definition}
The process of generating planning tasks can be formalized as a function $f: I \rightarrow (\mathbb{T},~\mathbb{E})$, where $I$ is the input space (e.g., instructions or prompts) and tuple $(\mathbb{T},~\mathbb{E})$ is the space of all possible planning tasks and environments.
Based on the definition in Section~\ref{sec:agent_task_definition}, we can express this as \( f(i) = (\mathbb{T}_i,~\mathbb{E}_i), \quad i \in I \), where $\mathbb{T}_i$ is the generated planning task and $\mathbb{E}_i$ is the generated environment for a given input $i$.
Our two-stage approach can be further decomposed as follows: 
i) \textbf{\textit{Environment Generation}} (\S\ref{sec:environment_generation}):
In the first stage, we generate the environment $\mathbb{E}_i$ based on the input instruction $i$. This can be represented as $\mathbb{E}_i = g_\mathbb{E}(i)$,  where $g_\mathbb{E}$ is the environment generation function that takes the instruction $i$ as input and produces the environment $\mathbb{E}_i$.
ii) \textbf{\textit{Task Generation}} (\S\ref{sec:goal_generation}):
In the second stage, we generate the task $\mathbb{T}_i$, conditioned on the environment $\mathbb{E}_i$ generated in the first stage. This can be expressed as:
$\mathbb{T}_i = g_\mathbb{T}(i, \mathbb{E}_i)$,
where $g_\mathbb{T}$ is the task generation function that takes both the original instruction $i$ and the generated environment $\mathbb{E}_i$ as inputs to produce the task $\mathbb{T}_i$.
We will detail the implementation of these two stages in the following section.

\subsection{Environment Generation}
\label{sec:environment_generation}

\begin{figure}[htbp]
    \centering
    \includegraphics[width=\linewidth]{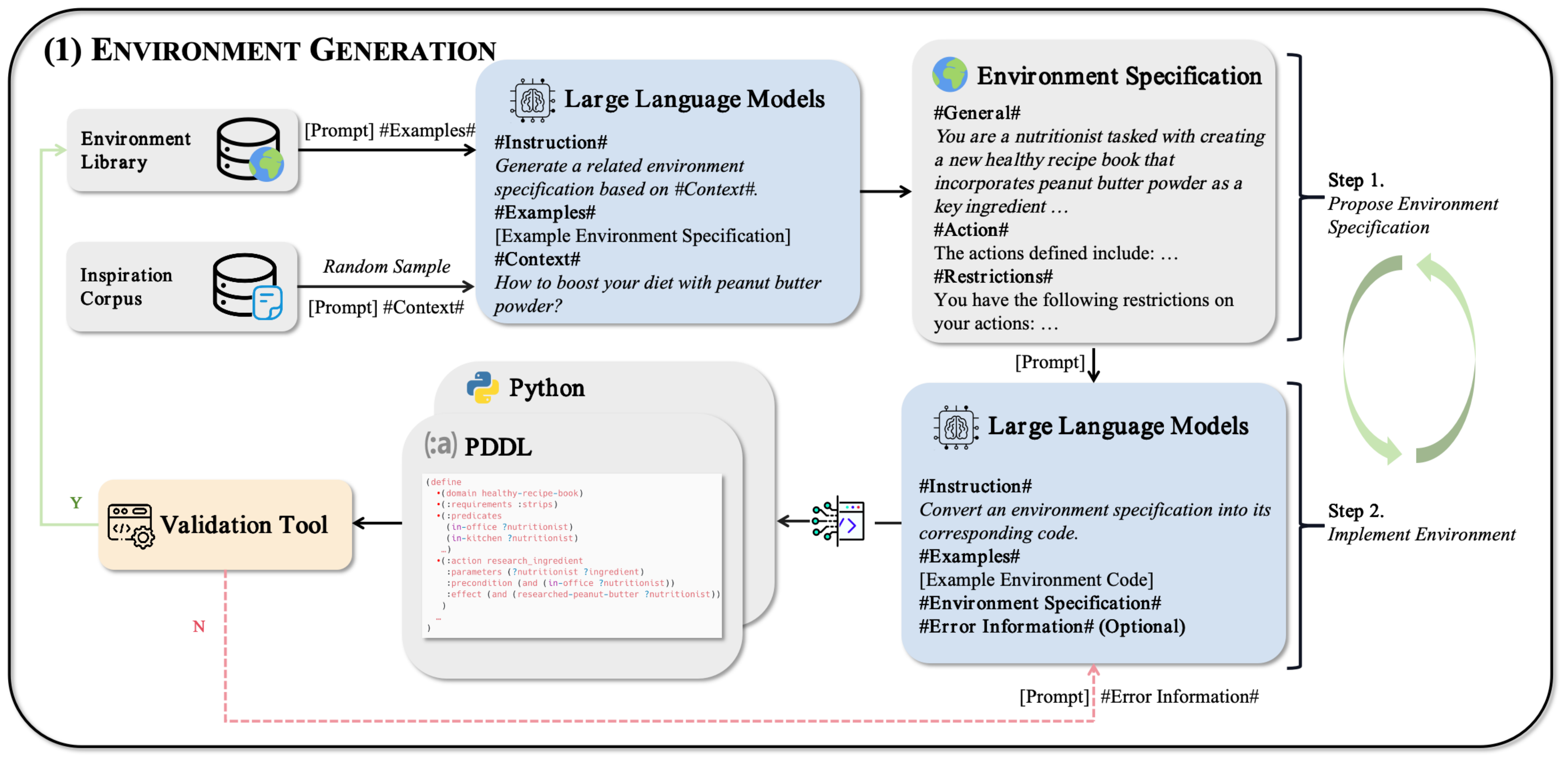}
    \caption{
    Overview of the process of environment generation. 
    }
    \label{fig:environment_generation}
\end{figure}

\paragraph{Overview}
As is shown in Figure~\ref{fig:environment_generation}, we propose a sophisticated framework for environment generation structured around three main components: (1) an \textit{environment specification generation} module where an LLM first generates a specification of the environment, typically including a general overview of the environment, descriptions of the state space and action space, and definitions of the transition functions; (2) an \textit{environment implementation} module that generates corresponding code based on the environment specification; and (3) an \textit{environment library} that stores previously generated high-quality environments, serving as a comprehensive environment dataset and providing in-context examples for generating new environments. Each component will be elaborated on in the following paragraph.

\paragraph{Environment Specification}
We initially prompt the LLM to generate an environment specification, which typically includes an overall depiction of the environment, specific actions and their corresponding preconditions and effects, and certain restrictions within the environment.
The environment specification will serve as the basis for generating specific environment codes. This two-stage approach, similar to the Chain-of-Thought~\cite{wei2022chain}, can better assist the LLM in creating high-quality environments.
For generating environment specifications, One direct approach is to prompt LLMs to generate random environments. 
However, due to the inherent inductive bias of LLMs, they struggle to generate diverse environments in this way.
Therefore, to address this issue, we build an inspiration corpus $D=\{t_0,t_1,\cdot,t_n\}$, containing sufficiently diverse text segments used to serve as the "inspiration" for generating environment specification with LLMs. 
More specifically, when generating an environment, we first sample a text segment $t_i$ from $D$, then prompt the LLM to generate a related environment based on $t_i$.
Taking the example in Figure~\ref{fig:environment_generation}, we first sample a text segment "\textit{How to boost your diet with peanut butter powder?}" from $D$. Then we prompt an LLM to generate a related environment where the agent is defined as a nutritionist tasked with creating a new healthy recipe book that prominently features peanut butter powder as a key ingredient. 
This approach significantly enhances the diversity of generated environments, thereby empowering more generalized agent training.
The inspiration corpus can be implemented in various ways, such as using a large-scale pre-trained corpus like Common Crawl. Alternatively, a domain-specific corpus, such as a code generation dataset~\cite{lai2023ds,chen2021evaluating}, can be used to generate environments for a specific domain. This paper uses LIMA~\cite{zhou2024lima} as the inspiration corpus, an instruction-tuning dataset with sufficient diversity.

\paragraph{Environment Implementation}
Conditioned on the generated environment specification, we generate its corresponding code, i.e., implementing the environment.
This can be formulated as a typical code-generation problem with LLMs.
We also introduce a validation tool capable of capturing syntax errors to provide feedback during the code generation process, thereby iteratively refining it.

\paragraph{Environment Library}
We define the library at iteration $t$ as: $L_t = L_0 \cup \bigcup_{k=1}^{t} \{\mathbb{E}_i | \mathbb{E}_i = g_{\mathbb{E}}(i, L_{k-1}), i \in I_k, v(\mathbb{E}_i) = true\}$,
where $L_0$ is the initial seed library, and the union represents all verified environments generated up to iteration $t$.
This iterative process allows continuous expansion and refinement of the environment library, potentially leading to increasingly complex and diverse environments over time.

\subsection{Task Generation}
\label{sec:goal_generation}

\begin{figure}[htbp]
    \centering
    \includegraphics[width=\linewidth]{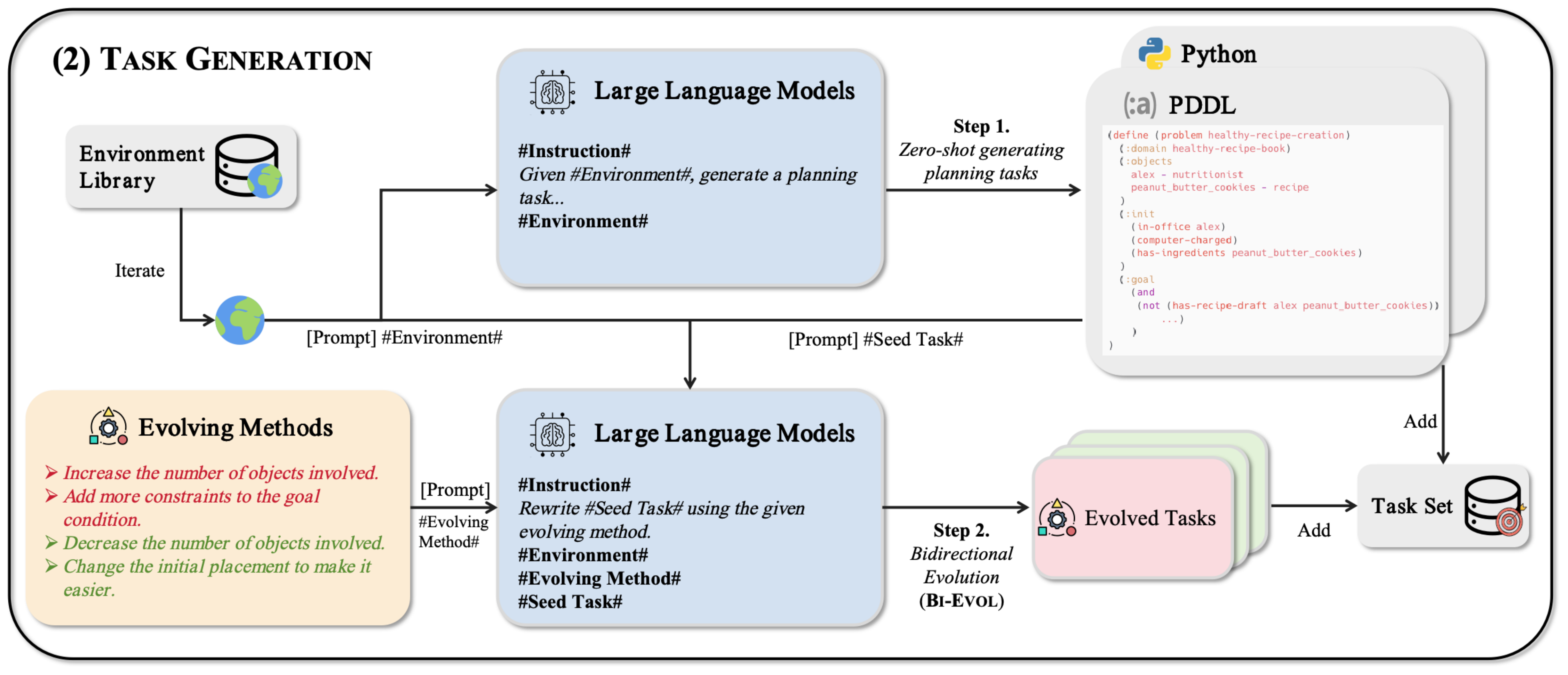}
    \caption{
    Overview of the process of task generation.
    The two-stage task generation process includes first generating unconditioned tasks, then applying \goal to evolve these planning tasks. Ultimately, both parts are incorporated into the task set. 
    In examples of evolving methods, red indicates evolution towards more difficult tasks, while green indicates the opposite.
    }
    \label{fig:goal_generation}
\end{figure}

\paragraph{Overview}
As depicted in Figure~\ref{fig:goal_generation}, conditioned on the generated environments, we prompt LLMs to generate corresponding planning tasks.
We employ a two-stage generation approach \goal for creating a diverse range of planning tasks in terms of difficulty. 
We begin by prompting the LLM with a specific environment, enabling it to generate an initial set of planning tasks in a zero-shot way.
Subsequently, we adjust these tasks to make them simpler or more challenging, forming a comprehensive set of planning tasks.

\paragraph{Bidirectional Evolution}
Many studies have proposed evolving instructions, primarily focusing on making instructions more difficult~\cite{xu2023wizardlm,luo2023wizardcoder,luo2023wizardmath}. The effectiveness of this approach relies heavily on the assumption that LLMs inherently possess the ability to follow simple instructions. 
However, according to findings from some studies~\cite{ma2024agentboard,liu2023agentbench}, LLMs often exhibit poor performance even in simple planning tasks. 
Therefore, we propose \textbf{\goal}, which introduces evolution in two directions: easy-evol and hard-evol. 
Easy-evol typically involves simplifying the goal conditions. The motivation is that easier tasks can facilitate learning when the agent performs poorly and cannot directly learn from typically difficult goals. 
Conversely, hard-evol usually involves making the goal conditions more complex, increasing the number of steps required for the agent to complete the task. This can further enhance the agent's capability to perform the planning task.
To our knowledge, we are the first to introduce bidirectional evolution in the agent data generation scenario.
The prompt examples are shown in Figure~\ref{fig:goal_generation}.

\section{Experiments}
\label{sec:experiments}
\vspace{-5pt}

To evaluate the effectiveness of the proposed framework, we synthesize environments and planning tasks using the Planning Domain Definition Language (PDDL), a widely adopted programming language for planning.
Subsequently, we evaluate its performance across various unseen planning tasks in a \textbf{zero-shot} manner. To validate the effectiveness and generalizability of \ours, we categorized the evaluated tasks into two distinct groups: 
i) \textit{In-Domain Tasks}: Planning tasks implemented using PDDL.
ii) \textit{Out-of-Domain Tasks}: These comprise tasks developed using other programming languages, such as Python.

\subsection{Experimental Setup}

\paragraph{Evaluation Tasks}
For \textit{In-Domain Tasks}, we select four widely used PDDL-based planning tasks: Blocksworld, Gripper, Tyreworld, and Barman~\cite{ma2024agentboard}.
More explicitly, Blocksworld requires an agent to achieve a target configuration by moving blocks, while Gripper involves moving objects between different rooms.
Tyreworld simulates changing a car tire, including removing the flat tire, replacing it with a spare, and installing the new tire. 
Barman emulates a bartender's tasks in mixing cocktails, which include combining various ingredients, using shakers, and garnishing drinks.
For \textit{Out-of-Domain Tasks}, we select three challenging partial-observable planning tasks: Alfworld~\cite{shridhar2020alfworld} and BabyAI~\cite{chevalier2018babyai}, Jericho~\cite{jericho}. 
Alfworld is an environment designed to test agents' abilities to perform everyday household tasks.
While in BabyAI, the agent interprets and executes natural language instructions in a grid-world setting.
Jericho~\cite{jericho} is a collection of text-based interactive fiction games in which players issue textual commands to alter the environment.

\paragraph{Evaluation Metrics}

We utilized two evaluation metrics to evaluate planning ability: \textit{success rate} and \textit{progress rate}~\cite{ma2024agentboard}. 
During each interaction round, we assigned a progress rate, denoted as \( r_t \), to measure the progression towards the goal state \( g \). 
As the agent transitions through states \( s_t = [s_0, \ldots, s_t] \), its progress is assessed using a matching score \( f(\cdot, g) \rightarrow [0, 1] \), which quantifies the similarity between the current state and the goal state. 
Initially, \( r_t \) is set to 0, indicating no progress. 
Only when the progress rate reaches 1 does the success rate attain 1; all other scenarios yield a 0 outcome. 
The success rate reflects the agent's capacity to complete a comprehensive task.

\paragraph{Baselines}

We compare \ours with a series of widely-used multipurpose foundation models that exhibit state-of-the-art performance, such as GPT-3.5~\cite{opeiai2022gpt} and GPT-4~\cite{openai2023gpt}, CodeLlama~\cite{roziere2023code}, Mistral~\cite{jiang2023mistral}, Llama-2~\cite{touvron2023llama}, and Llama-3.1~\cite{meta2024llama3}. We use their instruct-tuned versions for all multipurpose foundation models (\S\ref{app:models}). Additionally, some models have undergone specialized training on agent trajectory data, such as AgentLM~\cite{zeng2023agenttuning}, FireAct~\cite{chen2023fireact}, Agent-Flan~\cite{chen2024agent}.
We also utilize the AgentInstruct~\cite{zeng2023agenttuning} dataset to train Llama-3.1, following the training configuration of \ours as a baseline model.

\paragraph{Implementation Details}
We followed the environment and task implementation of AgentBoard~\cite{ma2024agentboard}. For the configuration of evaluation tasks, we employ act-only prompting~\cite{yao2022react}, setting the maximum step length for the LLM agent to 30.
We selected LIMA~\cite{zhou2024lima} as the text corpus $D$ for generating environments, which leverages various data manipulation techniques to ensure a diverse range of instructions. 
For environment generation and task generation, we employ GPT-4~\footnote{We applied the gpt-4-20230321 API from Azure OpenAI service.}, configuring the inference parameters with a temperature of 0 and a top\_p value of 0.95. 
Based on \ours, we generated a total of 592 environments. For each environment, we generated ten unconditioned tasks, which were then evolved into ten refined tasks using \goal. To generate trajectory data for training, we utilized the domain-independent planner FastDownward\footnote{\url{https://www.fast-downward.org/}}, ensuring optimal trajectory data. This process ultimately led to 7246 trajectories. 
More details of the dataset can be found in Appendix~\ref{app:stat} and~\ref{app:examples}.
Since the trajectory data is structured, such as \textit{"pickup(o1)"}, we employ GPT-4 to generate a natural language mapping, for example, \textit{"pick up object \{arg1\}"}, to transform structured actions into natural language actions. We detailed the generation of natural language mapping in~\ref{app:nl_map}.
During the training process, we employed Llama-3.1-8B (base version) as our foundation model, blending general instruction data from the ShareGPT dataset in a 1:4 ratio. For the 70B model, we selected Llama-3.1-70B-Instruct and trained it using LoRA~\cite{hu2021lora} without incorporating general instructions, with a rank of 16. The hyperparameters were configured as follows: a batch size of 64, 10 epochs, a context length of 4096 tokens, and no warmup steps. Checkpoints from epochs 5 through 10 were retained and subsequently evaluated on in-domain tasks. The model demonstrating optimal performance was then selected for further evaluation on out-of-domain tasks.
We conducted all experiments utilizing V100 and A100 GPUs.

\subsection{Evaluation on In-Domain Tasks}

\begin{table}[htbp]
\centering
\caption{Performance comparison between \ours and baseline models in in-domain tasks. \textit{``Overall''} is the weighted average of performance in different tasks. ``SR'' and ``PR'' stand for ``success rate'' and ``progress rate'' metrics.}
\vspace{5pt}
\label{tab:pddl-comparison}
\resizebox{0.95\textwidth}{!}{%
\begin{tabular}{l|cc|rrrrrrrrrr}
\toprule
\multirow{2}{*}{\textbf{Model}} & \multirow{2}{*}{\textbf{Size}} & \multirow{2}{*}{\textbf{Version}} & \multicolumn{2}{c}{\textbf{Gripper}} & \multicolumn{2}{c}{\textbf{Blockworld}} & \multicolumn{2}{c}{\textbf{Barman}} & \multicolumn{2}{c}{\textbf{Tyreworld}} & \multicolumn{2}{c}{\textbf{Overall}} \\
\cmidrule{4-13} & & & \textbf{SR} & \textbf{PR} & \textbf{SR} & \textbf{PR} & \textbf{SR} & \textbf{PR} & \textbf{SR} & \textbf{PR} & \textbf{SR} & \textbf{PR} \\
\midrule
\multirow{2}{*}{\textsc{GPT-4}} & \texttt{-} & {\small\texttt{2023-05-15}} & 55.0 & 83.3 & 50.0 & 75.0 & 75.0 & 82.5 & 60.0 & 80.3 & 61.7 & 81.2 \\
& \texttt{-} & {\small\texttt{turbo}} & 50.0 & 87.8 & 40.0 & 71.7 & 10.0 & 17.5 & 10.0 & 39.3 & 23.3 & 44.7 \\
\midrule
\multirow{2}{*}{\textsc{GPT-3.5}} & \texttt{-} & {\small\texttt{turbo}} & 0.0 & 30.6 & 0.0 & 18.3 & 10.0 & 21.7 & 10.0 & 27.1 & 5.0 & 25.0 \\
 & \texttt{-} & {\small\texttt{turbo-16k}} & 0.0 & 28.2 & 0.0 & 20.0 & 5.0 & 13.3 & 10.0 & 32.7 & 3.3 & 22.6 \\
\midrule
\multirow{3}{*}{\textsc{CodeLlama}} & 7B & {\small\texttt{instruct}} & 0.0 & 7.4 & 0.0 & 8.3 & 0.0 & 0.0 & 10.0 & 26.0 & 1.7 & 8.2 \\
& 13B & {\small\texttt{instruct}} & 5.0 & 15.6 & 0.0 & 5.0 & 0.0 & 0.0 & 0.0 & 19.3 & 1.7 & 9.3 \\
& 34B & {\small\texttt{instruct}} & 0.0 & 28.7 & 10.0 & 21.7 & 5.0 & 7.5 & 0.0 & 17.1 & 3.3 & 18.5 \\
\midrule
\textsc{Mistral} & 7B & {\small\texttt{instruct-v0.2}} & 0.0 & 5.3 & 0.0 & 10.0 & 0.0 & 2.5 & 0.0 & 7.3 & 0.0 & 5.5 \\
\midrule
\multirow{3}{*}{\textsc{Llama-2}} & 7B & {\small\texttt{chat}} & 0.0 & 1.5 & 0.0 & 0.0 & 0.0 & 0.0 & 0.0 & 0.0 & 0.0 & 0.5 \\
& 13B & {\small\texttt{chat}} & 0.0 & 0.0 & 0.0 & 6.7 & 0.0 & 1.7 & 0.0 & 14.8 & 0.0 & 4.1 \\
& 70B & {\small\texttt{chat}} & 0.0 & 8.8 & 0.0 & 5.0 & 5.0 & 9.2 & 0.0 & 7.8 & 1.7 & 8.1 \\
\midrule
\textsc{FireAct} & 7B & {\small\texttt{-}} & 0.0 & 0.0 & 0.0 & 5.0 & 0.0 & 0.0 & 0.0 & 0.0 & 0.0 & 1.5 \\
\midrule
\textsc{Agent-Flan} & 7B & {\small\texttt{-}} & 0.0 & 0.0 & 0.0 & 0.0 & 0.0 & 0.0 & 0.0 & 0.0 & 0.0 & 0.0 \\
\midrule
\multirow{2}{*}{\textsc{AgentLM}} & 7B & {\small\texttt{-}} & 0.0 & 0.0 & 0.0 & 0.0 & 0.0 & 2.5 & 0.0 & 0.0 & 0.0 & 0.8 \\
& 70B & {\small\texttt{-}} & 0.0 & 0.8 & 0.0 & 6.7 & 5.0 & 13.3 & 10.0 & 26.0 & 3.3 & 10.2 \\
\midrule
\textsc{Llama-3.1} & 8B & {\small\texttt{instruct}} & 0.0 & 0.0 & 0.0 & 1.7 & 0.0 & 0.0 & 0.0 & 16.2 & 0 & 3.0 \\
~$w.$ \textsc{AgentInstruct} & 8B & {\small\texttt{-}} & 0.0 & 3.4 & 0.0 & 5.0 & 0.0 & 6.7 & 10.0 & 26.0 & 1.7 & 8.5 \\
\rowcolor{blue!15}
~$w.$ \ours & 8B & {\small\texttt{-}} & 20.0 & 45.2 & 20.0 & 31.7 & 10.0 & 32.7 & 10.0 & 32.7 & 15.0 & 33.3 \\
\midrule
\textsc{Llama-3.1} & 70B & {\small\texttt{instruct}} & 55.0 & 89.3 & 50.0 & 70.0 & 70.0 & 80.0 & 40.0 & 65.3 & 56.7 & 79.0 \\
~$w.$ \textsc{AgentInstruct} & 70B & {\small\texttt{-}} & 45.0 & 78.9 & 70.0 & 80.0 & 25.0 & 32.5 & 50.0 & 74.4 & 43.4 & 62.9 \\
\rowcolor{blue!15}
~$w.$ \ours & 70B & {\small\texttt{-}} & 55.0 & 89.3 & 50.0 & 63.3 & 70.0 & 82.5 & 50.0 & 82.2 & 58.3 & 81.5 \\
\bottomrule
\end{tabular}%
}
\end{table}

As shown in Table~\ref{tab:pddl-comparison}, the \ours-tuned Llama-3.1-8B model outperforms GPT-3.5 in overall progress rate (33.3 vs. 25.0). Furthermore, the \ours-tuned Llama-3.1-70B model slightly surpasses GPT-4 (81.5 vs. 81.2). 
When compared to other models with similar parameter scales, \ours consistently demonstrates superior performance across four distinct tasks. In relation to the base Llama-3.1 model, our model exhibits a substantial improvement for both the 8B and 70B versions, with overall progress rates increasing by 30.3 and 2.5, respectively. Notably, in tasks where the success rate of Llama-3.1-8B is zero, \ours achieves significant breakthroughs, further validating the efficacy of \ours.
From the above, we can draw the following conclusions:
\textit{i)} \ours-tuned Llama-3.1-8B outperforms GPT-3.5 in overall performance, while the 70B version achieves state-of-the-art results;
\textit{ii)} \ours-tuned Llama-3.1 has significantly improved both success rate and progress rate;
\textit{iii)} \ours consistently outperforms other models with similar parameter scales.

\subsection{Robustness}


\begin{table}[htbp]
    \small
  \centering
  \caption{Overall performance comparison of models before and after training with \ours on in-domain tasks. ``SR'' and ``PR'' stands for ``success rate'' and ``progress rate'' respectively.}
    \begin{tabular}{crrrrrr}
    \toprule
    \multirow{2}[4]{*}{\textbf{Model}} & \multicolumn{2}{c}{\textbf{Before}} & \multicolumn{2}{c}{\textbf{After}} & \multicolumn{2}{c}{\textbf{$\Delta$}} \\
\cmidrule{2-7}          & \textbf{SR}    & \textbf{PR}    & \textbf{SR}    & \textbf{PR}    & \textbf{SR}    & \textbf{PR} \\
    \midrule
    Llama-3-8B & 1.7  & 13.4 & 11.7 & 23.0 & \cellcolor[rgb]{ .973,  .412,  .42} 10.0 & \cellcolor[rgb]{ .976,  .443,  .451} 9.6 \\
    CodeLlama-7B & 1.7  & 8.2  & 6.7  & 18.1 & \cellcolor[rgb]{ .984,  .761,  .769} 5.0 & \cellcolor[rgb]{ .976,  .42,  .427} 9.9 \\
    Mistral-7B & 0.0     & 5.5  & 1.7  & 10.4 & \cellcolor[rgb]{ .988,  .988,  1} 1.7 & \cellcolor[rgb]{ .984,  .765,  .776} 4.9 \\
    \bottomrule
    \end{tabular}%
  \label{tab:performance_comparison}
\end{table}%

To validate the robustness of the constructed dataset with \ours, we conducted a series of experiments to evaluate its performance across different foundation models. 
We selected several widely used 7-8B foundation models, including Llama-3-8B, CodeLLama-7B, and Mistral-7B, to test the versatility and effectiveness of \ours.
As is shown in Table~\ref{tab:performance_comparison}, all three models exhibited significant improvements after training, with Llama-3-8B showing the highest success rate increase of 10.0 and CodeLlama-7B demonstrating a maximum progress rate increase of 9.9. These experimental results prove that the dataset constructed with \ours for agent training is highly effective across different models.

\subsection{Evaluation on Out-of-Domain Tasks}

\begin{table}[htbp]
\centering
\caption{\small{Performance comparison between \ours and baseline models on out-of-domain tasks. ``SR'' and ``PR'' stand for ``success rate'' and ``progress rate'' metrics.}}
\vspace{5pt}
\label{tab:other-tasks}
\resizebox{\linewidth}{!}{%
\begin{tabular}{l|cc|rrrrrr|rr}
\toprule
\multirow{2}{*}{\textbf{Model}} & \multirow{2}{*}{\textbf{Size}} & \multirow{2}{*}{\textbf{Version}} & \multicolumn{2}{c}{\textbf{Alfworld~\cite{shridhar2020alfworld}}} & \multicolumn{2}{c}{\textbf{BabyAI}~\cite{chevalier2018babyai}} & \multicolumn{2}{c}{\textbf{Jericho}~\cite{jericho}} & \multicolumn{2}{c}{\textbf{Average}} \\

\cmidrule{4-11} & & & \textbf{SR} & \textbf{PR} & \textbf{SR} & \textbf{PR} & \textbf{SR} & \textbf{PR} & \textbf{SR} & \textbf{PR} \\
\midrule
\textsc{GPT-4} & \texttt{-} & {\small\texttt{2023-05-15}} & 43.4 & 65.5 & 56.2 & 70.7 & 35.0 & 52.4 & 44.9 & 62.9\\
\midrule
\multirow{2}{*}{\textsc{GPT-3.5}} & \texttt{-} & {\small\texttt{turbo}} & 17.2 & 35.6 & 18.9 & 31.9 & 0.0 & 20.4 & 12.0 & 29.3 \\
& \texttt{-} & {\small\texttt{turbo-16k}} & 4.5 & 25.2 & 33.9 & 45.1 & 0.0 & 16.1 & 12.8 & 28.8 \\

\midrule
\multirow{3}{*}{\textsc{CodeLlama}} & 7B & {\small\texttt{instruct}} & 1.4 & 2.2 & 15.2 & 28.3 & 0.0 & 9.2 & 5.5 & 13.9 \\
& 13B & {\small\texttt{instruct}} & 2.2 & 13.4 & 17.0 & 22.2 & 0.0 & 0.0 & 6.4 & 11.9 \\
& 34B & {\small\texttt{instruct}} & 3.0 & 11.3 & 13.4 & 19.9 & 0.0 & 15.5 & 5.5 & 15.6 \\
\midrule
\textsc{Mistral} & 7B & {\small\texttt{instruct-v0.2}} & 0.0 & 9.8 & 18.1 & 24.4 & 0.0 & 12.1 & 6.0 & 15.4 \\
\midrule
\multirow{3}{*}{\textsc{Llama-2}} & 7B & {\small\texttt{chat}} & 0.0 & 1.5 & 5.4 & 8.3 & 0.0 & 4.2 & 1.8 & 4.7 \\
& 13B & {\small\texttt{chat}} & 0.0 & 7.8 & 6.2 & 18.2 & 0.0 & 3.2 & 2.1 & 9.7 \\
& 70B & {\small\texttt{chat}} & 3.0 & 13.2 & 19.6 & 30.0 & 0.0 & 7.8 & 7.5 & 17.0 \\
\midrule
\textsc{FireAct} & 7B & {\small\texttt{-}} & 0.0 & 0.8 & 4.5 & 8.6 & 0.0 & 2.8 & 1.5 & 4.7 \\
\midrule
\textsc{Agent-Flan} & 7B & {\small\texttt{-}} & 0.0 & 0.8 & 0.0 & 0.0 & 0.0 & 0.0 & 0.0 & 0.3 \\
\midrule
\multirow{2}{*}{\textsc{AgentLM}}$\dagger$ & 7B & {\small\texttt{-}} & {\small\texttt{-}} & {\small\texttt{-}} & 8.0 & 9.9 & 5.5 & 15.2 & {\small\texttt{-}} & {\small\texttt{-}} \\
& 70B & {\small\texttt{-}} & {\small\texttt{-}} & {\small\texttt{-}} & 27.7 & 37.1 & 0.0 & 18.4 & {\small\texttt{-}} & {\small\texttt{-}} \\
\midrule
\textsc{Llama-3.1} & 8B & {\small\texttt{instruct}} & 0.0 & 10.5 & 17.9 & 33.6 & 0.0 & 8.8 & 6.0 & 17.6 \\
~$w.$ \textsc{AgentTuning}$\dagger$ & 8B & {\small\texttt{-}} & {\small\texttt{-}} & {\small\texttt{-}} & 10.7 & 19.3 & 0.0 & 8.2 & {\small\texttt{-}} & {\small\texttt{-}} \\
\rowcolor{blue!15}
~$w.$ \ours & 8B & {\small\texttt{-}} & 17.9 & 31.7 & 32.1 & 46.0 & 0.0 & 14.3 & 16.0 & 30.7 \\
\midrule
\textsc{Llama-3.1} & 70B & {\small\texttt{instruct}} & 17.2 & 42.7 & 38.4 & 57.2 & 10.0 & 31.5 & 21.8 & 43.8 \\
~$w.$ \textsc{AgentTuning}$\dagger$ & 70B & {\small\texttt{-}} & {\small\texttt{-}} & {\small\texttt{-}} & 17.9 & 35.9 & 10.0 & 31.9 & {\small\texttt{-}} & {\small\texttt{-}} \\
\rowcolor{blue!15}
~$w.$ \ours & 70B & {\small\texttt{-}} & 19.4 & 46.1 & 42.0 & 62.2 & 15.0 & 38.1 & 25.5 & 48.8 \\
\bottomrule
\end{tabular}%
}
\end{table}


We also conducted evaluations on out-of-domain agent tasks. As illustrated in Table~\ref{tab:other-tasks}, similar experimental phenomena were observed. 
Firstly, \ours demonstrates a substantial performance improvement over Llama-3.1, with an increase of 13.1\% in the average progress rate for the 8B model and 5.0\% for the 70B model. Additionally, the \ours-tuned Llama-3.1-8B model outperforms GPT-3.5.
When compared to general models and agent fine-tuning models with similar parameter scales, \ours consistently outperforms them on both tasks. The superior performance on out-of-domain tasks further emphasizes the effectiveness and generalization capability of our data synthesis methods.

\section{Related Work}

\paragraph{Large Language Model based Agent.}

Large Language Models have demonstrated exceptional reasoning capabilities~\cite{touvron2023llama,meta2024llama3,opeiai2022gpt,openai2023gpt4,jiang2023mistral}. Owing to such abilities, over the past two years, LLM-based agents have experienced significant development~\cite{shinn2023reflexion,wu2023autogen,hong2023metagpt,sumers2023cognitive,wang2023survey,xi2023rise}.
Unlike the traditional method of using LLMs for text-based reasoning, such as Chain-of-Thought~\cite{wei2022chain}, LLM-based agents typically involve interaction with the environment, adjusting the output in a closed-loop manner based on environmental information.
These LLM-based agents, now fortified with capabilities like Memorizing~\cite{zhong2024memorybank,liu2023think,liang2023unleashing,yao2023retroformer,shinn2023reflexion,zhao2024expel,zhu2023ghost,tack2024online,huang2023recommender}, Tool-use~\cite{cheng2022binding,parisi2022talm,YongliangShen2023_HuggingGPT,li2023api,TimoSchick2023_Toolformer,qin2023toolllm}, and Planning~\cite{gao2024dag,brohan2023can,mu2024embodiedgpt,mu2024robocodex,ruan2023tptu,ajay2024compositional}, exhibit a marked enhancement in their overall efficacy. 
Although this paper mainly focuses on the planning capability of LLM-based agents, we believe \ours has the potential to generalize to other scenarios of LLM-based agents.

\paragraph{Planning with Large Language Models.}

Planning is one of the key applications of LLM-based agents, applicable in various scenarios such as robotic planning~\cite{shridhar2020alfworld,puig2018virtualhome,huang2022language,valmeekam2024planbench,ding2023task,wu2023embodied,liu2023llm+,guan2023leveraging}, travel planning~\cite{xie2024travelplanner,aghzal2023can}, calendar scheduling~\cite{zheng2024natural}, code generation~\cite{bairi2024codeplan} and others~\cite{wang2023describe}. 
It is typically defined as the process of systematically determining a sequence of actions or steps required to achieve a desired goal from an initial state, considering constraints and available resources. 
This definition primarily differentiates from studies that utilize LLMs to generate ungrounded plans as guidance for problem-solving~\cite{zhou2022least,wang2023plan}, rather than directly producing executable actions.
Planning can be categorized into two types: open-loop planning, where the LLM outputs an entire action sequence before execution~\cite{huang2022language,valmeekam2024planbench}, and closed-loop planning, where the LLM-based agent decides the next action based on real-time environmental interaction after executing a previous action~\cite{singh2023progprompt,brohan2023can,sun2023adaplanner,sun2023pearl,lin2023grounded,song2023llmplanner,huang2023grounded,huang2022inner}. 
This paper mainly focuses on close-loop planning, which is more adaptable for error correction, human interaction, and environmental grounding. 
Recent studies on close-loop planning have integrated chain-of-thought reasoning into the planning process~\cite{yao2022react}. 
Additionally, some papers have explored the use of tree-search methods to enhance the performance of LLM planning~\cite{hu2023tree,hao2023reasoning,yao2023tree,liu2024tool,zhao2024large,wang2023promptagent,zhou2023language}.
Instead of designing novel frameworks or engaging in prompt engineering, this paper explores how training can enhance the planning capabilities of LLM-based agents.

\footnotetext{$\dagger$AgentTuning utilized Alfworld's training set, meaning Alfworld cannot be considered an out-of-domain task. Consequently, we did not evaluate the performance of AgentLM or the AgentTuning-trained model on Alfworld.}

\paragraph{Agent Training.}

Recently, numerous studies have aimed to enhance LLM-based agent capabilities by incorporating agent trajectory data into their training~\cite{wang2024learning,chen2024agent,zhang2024agentohana,wang2024llms,song2024trial}. 
Advanced works such as AgentTuning~\cite{zeng2023agenttuning} utilize GPT-4 to generate trajectory data across six distinct environments. Subsequently, this data is filtered and employed in training Large Language Models, enhancing the agent capabilities of base models.
Another work, FireAct~\cite{chen2023fireact}, proposes training with both CoT data and ReAct format data, enabling the model to discern when to use reasoning to solve problems and when to call external tools. 
Agent LUMOS~\cite{yin2023lumos} suggests separately training Planning and Grounding models, enabling LLM-based agents to learn to decompose complex problems before execution.
LLM-Modulo framwork~\cite{kambhampati2024llms} proposes to leverage LLMs generating candidate plans and verify them with an external verifier. Then, use the verified trajectories for fine-tuning LLMs.
Similarly, ~\cite{arora2023learning} takes a generate-test loop to synthesize trajectories for LLM training.
Unlike previous papers on all agent training, \ours goes beyond merely generating trajectory data using Large Language Models. Instead, we utilize Large Language Models to generate agent environments, which can be considered a more foundational application. As a result, we have constructed over 500 environments for training, whereas previous works typically use fewer than 10 environments to synthesize agent data.

\paragraph{Environment and Task Generation with Large Language Models.}

The utilization of LLMs to generate environments and tasks is an emerging application.
Some studies have explored utilizing LLMs to generate layouts in robotic simulations, typically involving the creation of configuration files~\cite{wang2023robogen,yang2024holodeck,wang2023gensim}. 
While these methods can construct numerous scene-level environments, they often struggle to achieve diversity at the underlying mechanism level. 
AgentTuning~\cite{zeng2023agenttuning} employs a task generation approach similar to the Self-instruct~\cite{wei2021finetuned} method, using the test set as seed data. This approach not only poses a risk of data leakage but also leads to insufficient diversity in task difficulty.
ByteSized32~\cite{wang2023bytesized32} uses LLMs to generate Python-based games based on predefined task specifications automatically. 
Similarly, other works~\cite{guan2023leveraging} leverage LLMs to automatically construct PDDL domains based on a task specification.
In contrast to these studies, this paper proposes using a diverse text corpus to generate environment code automatically. 
This approach facilitates the creation of a wide range of rich environments without predefined definitions.

\section{Conclusion}
\vspace{-5pt}

In this paper, we explore using LLMs to automatically generate environment and planning tasks for LLM-based agent training. 
Specifically, for generating diverse environments, we propose utilizing an inspiration corpus composed of various domain-specific text segments as the context for environment synthesis. To enhance the difficulty diversity of generated planning tasks, we introduce a bidirectional evolution method, \goal, which evolves planning tasks from both easier and more challenging directions to create a task set with a more gradual difficulty curve, thereby improving the effectiveness of LLM learning. Based on \ours, we developed a dataset consisting of 592 environments and 7246 trajectories and trained it on a series of LLMs. 
The \ours-tuned Llama-3.1-8B model surpassed GPT-3.5 on planning tasks, while the \ours-tuned Llama-3.1-70B model achieved a new state-of-the-art performance.

\clearpage
\bibliographystyle{plainnat}
\bibliography{bibs/agent_benchmark,bibs/agent_prompting,bibs/agent_training,bibs/foundation_model,bibs/misc,bibliography,bibs/environment_generation}

\begin{thebibliography}{95}
\providecommand{\natexlab}[1]{#1}
\providecommand{\url}[1]{\texttt{#1}}
\expandafter\ifx\csname urlstyle\endcsname\relax
  \providecommand{\doi}[1]{doi: #1}\else
  \providecommand{\doi}{doi: \begingroup \urlstyle{rm}\Url}\fi

\bibitem[Aghzal et~al.(2023)Aghzal, Plaku, and Yao]{aghzal2023can}
Mohamed Aghzal, Erion Plaku, and Ziyu Yao.
\newblock Can large language models be good path planners? a benchmark and investigation on spatial-temporal reasoning.
\newblock \emph{arXiv preprint arXiv:2310.03249}, 2023.

\bibitem[Ajay et~al.(2024)Ajay, Han, Du, Li, Gupta, Jaakkola, Tenenbaum, Kaelbling, Srivastava, and Agrawal]{ajay2024compositional}
Anurag Ajay, Seungwook Han, Yilun Du, Shuang Li, Abhi Gupta, Tommi Jaakkola, Josh Tenenbaum, Leslie Kaelbling, Akash Srivastava, and Pulkit Agrawal.
\newblock Compositional foundation models for hierarchical planning.
\newblock \emph{Advances in Neural Information Processing Systems}, 36, 2024.

\bibitem[Arora and Kambhampati(2023)]{arora2023learning}
Daman Arora and Subbarao Kambhampati.
\newblock Learning and leveraging verifiers to improve planning capabilities of pre-trained language models.
\newblock \emph{arXiv preprint arXiv:2305.17077}, 2023.

\bibitem[Bairi et~al.(2024)Bairi, Sonwane, Kanade, Iyer, Parthasarathy, Rajamani, Ashok, and Shet]{bairi2024codeplan}
Ramakrishna Bairi, Atharv Sonwane, Aditya Kanade, Arun Iyer, Suresh Parthasarathy, Sriram Rajamani, B~Ashok, and Shashank Shet.
\newblock Codeplan: Repository-level coding using llms and planning.
\newblock \emph{Proceedings of the ACM on Software Engineering}, 1\penalty0 (FSE):\penalty0 675--698, 2024.

\bibitem[Brohan et~al.(2023)Brohan, Chebotar, Finn, Hausman, Herzog, Ho, Ibarz, Irpan, Jang, Julian, et~al.]{brohan2023can}
Anthony Brohan, Yevgen Chebotar, Chelsea Finn, Karol Hausman, Alexander Herzog, Daniel Ho, Julian Ibarz, Alex Irpan, Eric Jang, Ryan Julian, et~al.
\newblock Do as i can, not as i say: Grounding language in robotic affordances.
\newblock In \emph{Conference on robot learning}, pages 287--318. PMLR, 2023.

\bibitem[Chen et~al.(2023)Chen, Shu, Shareghi, Collier, Narasimhan, and Yao]{chen2023fireact}
Baian Chen, Chang Shu, Ehsan Shareghi, Nigel Collier, Karthik Narasimhan, and Shunyu Yao.
\newblock Fireact: Toward language agent fine-tuning.
\newblock \emph{arXiv preprint arXiv:2310.05915}, 2023.

\bibitem[Chen et~al.(2021)Chen, Tworek, Jun, Yuan, Pinto, Kaplan, Edwards, Burda, Joseph, Brockman, et~al.]{chen2021evaluating}
Mark Chen, Jerry Tworek, Heewoo Jun, Qiming Yuan, Henrique Ponde De~Oliveira Pinto, Jared Kaplan, Harri Edwards, Yuri Burda, Nicholas Joseph, Greg Brockman, et~al.
\newblock Evaluating large language models trained on code.
\newblock \emph{arXiv preprint arXiv:2107.03374}, 2021.

\bibitem[Chen et~al.(2024)Chen, Liu, Wang, Zhang, Liu, Lin, Chen, and Zhao]{chen2024agent}
Zehui Chen, Kuikun Liu, Qiuchen Wang, Wenwei Zhang, Jiangning Liu, Dahua Lin, Kai Chen, and Feng Zhao.
\newblock Agent-flan: Designing data and methods of effective agent tuning for large language models.
\newblock \emph{arXiv preprint arXiv:2403.12881}, 2024.

\bibitem[Cheng et~al.(2022)Cheng, Xie, Shi, Li, Nadkarni, Hu, Xiong, Radev, Ostendorf, Zettlemoyer, et~al.]{cheng2022binding}
Zhoujun Cheng, Tianbao Xie, Peng Shi, Chengzu Li, Rahul Nadkarni, Yushi Hu, Caiming Xiong, Dragomir Radev, Mari Ostendorf, Luke Zettlemoyer, et~al.
\newblock Binding language models in symbolic languages.
\newblock \emph{arXiv preprint arXiv:2210.02875}, 2022.

\bibitem[Chevalier-Boisvert et~al.(2018)Chevalier-Boisvert, Bahdanau, Lahlou, Willems, Saharia, Nguyen, and Bengio]{chevalier2018babyai}
Maxime Chevalier-Boisvert, Dzmitry Bahdanau, Salem Lahlou, Lucas Willems, Chitwan Saharia, Thien~Huu Nguyen, and Yoshua Bengio.
\newblock Babyai: A platform to study the sample efficiency of grounded language learning.
\newblock \emph{arXiv preprint arXiv:1810.08272}, 2018.

\bibitem[Ding et~al.(2023)Ding, Zhang, Paxton, and Zhang]{ding2023task}
Yan Ding, Xiaohan Zhang, Chris Paxton, and Shiqi Zhang.
\newblock Task and motion planning with large language models for object rearrangement, 2023.

\bibitem[Gao et~al.(2024)Gao, Mu, Qu, Hu, Guo, Luo, and Lu]{gao2024dag}
Zeyu Gao, Yao Mu, Jinye Qu, Mengkang Hu, Lingyue Guo, Ping Luo, and Yanfeng Lu.
\newblock Dag-plan: Generating directed acyclic dependency graphs for dual-arm cooperative planning.
\newblock \emph{arXiv preprint arXiv:2406.09953}, 2024.

\bibitem[Guan et~al.(2023)Guan, Valmeekam, Sreedharan, and Kambhampati]{guan2023leveraging}
Lin Guan, Karthik Valmeekam, Sarath Sreedharan, and Subbarao Kambhampati.
\newblock Leveraging pre-trained large language models to construct and utilize world models for model-based task planning.
\newblock \emph{Advances in Neural Information Processing Systems}, 36:\penalty0 79081--79094, 2023.

\bibitem[Hao et~al.(2023)Hao, Gu, Ma, Hong, Wang, Wang, and Hu]{hao2023reasoning}
Shibo Hao, Yi~Gu, Haodi Ma, Joshua~Jiahua Hong, Zhen Wang, Daisy~Zhe Wang, and Zhiting Hu.
\newblock Reasoning with language model is planning with world model.
\newblock \emph{arXiv preprint arXiv:2305.14992}, 2023.

\bibitem[Hausknecht et~al.(2020)Hausknecht, Ammanabrolu, C{\^o}t{\'e}, and Yuan]{jericho}
Matthew Hausknecht, Prithviraj Ammanabrolu, Marc-Alexandre C{\^o}t{\'e}, and Xingdi Yuan.
\newblock Interactive fiction games: A colossal adventure.
\newblock In \emph{Proceedings of the AAAI Conference on Artificial Intelligence}, volume~34, pages 7903--7910, 2020.

\bibitem[Hong et~al.(2023)Hong, Zheng, Chen, Cheng, Wang, Zhang, Wang, Yau, Lin, Zhou, et~al.]{hong2023metagpt}
Sirui Hong, Xiawu Zheng, Jonathan Chen, Yuheng Cheng, Jinlin Wang, Ceyao Zhang, Zili Wang, Steven Ka~Shing Yau, Zijuan Lin, Liyang Zhou, et~al.
\newblock {MetaGPT}: Meta programming for multi-agent collaborative framework.
\newblock \emph{arXiv preprint arXiv:2308.00352}, 2023.

\bibitem[Hu et~al.(2021)Hu, Shen, Wallis, Allen-Zhu, Li, Wang, Wang, and Chen]{hu2021lora}
Edward~J Hu, Yelong Shen, Phillip Wallis, Zeyuan Allen-Zhu, Yuanzhi Li, Shean Wang, Lu~Wang, and Weizhu Chen.
\newblock Lora: Low-rank adaptation of large language models.
\newblock \emph{arXiv preprint arXiv:2106.09685}, 2021.

\bibitem[Hu et~al.(2023)Hu, Mu, Yu, Ding, Wu, Shao, Chen, Wang, Qiao, and Luo]{hu2023tree}
Mengkang Hu, Yao Mu, Xinmiao Yu, Mingyu Ding, Shiguang Wu, Wenqi Shao, Qiguang Chen, Bin Wang, Yu~Qiao, and Ping Luo.
\newblock Tree-planner: Efficient close-loop task planning with large language models.
\newblock \emph{arXiv preprint arXiv:2310.08582}, 2023.

\bibitem[Huang et~al.(2022{\natexlab{a}})Huang, Abbeel, Pathak, and Mordatch]{huang2022language}
Wenlong Huang, Pieter Abbeel, Deepak Pathak, and Igor Mordatch.
\newblock Language models as zero-shot planners: Extracting actionable knowledge for embodied agents.
\newblock In \emph{International conference on machine learning}, pages 9118--9147. PMLR, 2022{\natexlab{a}}.

\bibitem[Huang et~al.(2022{\natexlab{b}})Huang, Xia, Xiao, Chan, Liang, Florence, Zeng, Tompson, Mordatch, Chebotar, Sermanet, Brown, Jackson, Luu, Levine, Hausman, and Ichter]{huang2022inner}
Wenlong Huang, Fei Xia, Ted Xiao, Harris Chan, Jacky Liang, Pete Florence, Andy Zeng, Jonathan Tompson, Igor Mordatch, Yevgen Chebotar, Pierre Sermanet, Noah Brown, Tomas Jackson, Linda Luu, Sergey Levine, Karol Hausman, and Brian Ichter.
\newblock Inner monologue: Embodied reasoning through planning with language models, 2022{\natexlab{b}}.

\bibitem[Huang et~al.(2023{\natexlab{a}})Huang, Xia, Shah, Driess, Zeng, Lu, Florence, Mordatch, Levine, Hausman, and Ichter]{huang2023grounded}
Wenlong Huang, Fei Xia, Dhruv Shah, Danny Driess, Andy Zeng, Yao Lu, Pete Florence, Igor Mordatch, Sergey Levine, Karol Hausman, and Brian Ichter.
\newblock Grounded decoding: Guiding text generation with grounded models for robot control, 2023{\natexlab{a}}.

\bibitem[Huang et~al.(2023{\natexlab{b}})Huang, Lian, Lei, Yao, Lian, and Xie]{huang2023recommender}
Xu~Huang, Jianxun Lian, Yuxuan Lei, Jing Yao, Defu Lian, and Xing Xie.
\newblock Recommender ai agent: Integrating large language models for interactive recommendations.
\newblock \emph{arXiv preprint arXiv:2308.16505}, 2023{\natexlab{b}}.

\bibitem[Hussein et~al.(2017)Hussein, Gaber, Elyan, and Jayne]{hussein2017imitation}
Ahmed Hussein, Mohamed~Medhat Gaber, Eyad Elyan, and Chrisina Jayne.
\newblock Imitation learning: A survey of learning methods.
\newblock \emph{ACM Computing Surveys (CSUR)}, 50\penalty0 (2):\penalty0 1--35, 2017.

\bibitem[Jiang et~al.(2023)Jiang, Sablayrolles, Mensch, Bamford, Chaplot, Casas, Bressand, Lengyel, Lample, Saulnier, et~al.]{jiang2023mistral}
Albert~Q Jiang, Alexandre Sablayrolles, Arthur Mensch, Chris Bamford, Devendra~Singh Chaplot, Diego de~las Casas, Florian Bressand, Gianna Lengyel, Guillaume Lample, Lucile Saulnier, et~al.
\newblock Mistral 7b.
\newblock \emph{arXiv preprint arXiv:2310.06825}, 2023.

\bibitem[Kaelbling and Lozano-Pérez(2011)]{kaelblingTAMP}
Leslie~Pack Kaelbling and Tomás Lozano-Pérez.
\newblock Hierarchical task and motion planning in the now.
\newblock In \emph{2011 IEEE International Conference on Robotics and Automation}, pages 1470--1477, 2011.
\newblock \doi{10.1109/ICRA.2011.5980391}.

\bibitem[Kambhampati et~al.(2024)Kambhampati, Valmeekam, Guan, Stechly, Verma, Bhambri, Saldyt, and Murthy]{kambhampati2024llms}
Subbarao Kambhampati, Karthik Valmeekam, Lin Guan, Kaya Stechly, Mudit Verma, Siddhant Bhambri, Lucas Saldyt, and Anil Murthy.
\newblock Llms can't plan, but can help planning in llm-modulo frameworks.
\newblock \emph{arXiv preprint arXiv:2402.01817}, 2024.

\bibitem[Lai et~al.(2023)Lai, Li, Wang, Zhang, Zhong, Zettlemoyer, Yih, Fried, Wang, and Yu]{lai2023ds}
Yuhang Lai, Chengxi Li, Yiming Wang, Tianyi Zhang, Ruiqi Zhong, Luke Zettlemoyer, Wen-tau Yih, Daniel Fried, Sida Wang, and Tao Yu.
\newblock Ds-1000: A natural and reliable benchmark for data science code generation.
\newblock In \emph{International Conference on Machine Learning}, pages 18319--18345. PMLR, 2023.

\bibitem[Li et~al.(2023)Li, Zhao, Yu, Song, Li, Yu, Li, Huang, and Li]{li2023api}
Minghao Li, Yingxiu Zhao, Bowen Yu, Feifan Song, Hangyu Li, Haiyang Yu, Zhoujun Li, Fei Huang, and Yongbin Li.
\newblock Api-bank: A comprehensive benchmark for tool-augmented llms.
\newblock \emph{arXiv preprint arXiv:2304.08244}, 2023.

\bibitem[Liang et~al.(2023)Liang, Wang, Huang, Wu, Wu, Lu, Ma, and Li]{liang2023unleashing}
Xinnian Liang, Bing Wang, Hui Huang, Shuangzhi Wu, Peihao Wu, Lu~Lu, Zejun Ma, and Zhoujun Li.
\newblock Unleashing infinite-length input capacity for large-scale language models with self-controlled memory system.
\newblock \emph{arXiv e-prints}, pages arXiv--2304, 2023.

\bibitem[Lin et~al.(2023)Lin, Huang, Liu, Gu, Sommerer, and Ren]{lin2023grounded}
Bill~Yuchen Lin, Chengsong Huang, Qian Liu, Wenda Gu, Sam Sommerer, and Xiang Ren.
\newblock On grounded planning for embodied tasks with language models.
\newblock In \emph{Proceedings of the AAAI Conference on Artificial Intelligence}, volume~37, pages 13192--13200, 2023.

\bibitem[Liu et~al.(2023{\natexlab{a}})Liu, Jiang, Zhang, Liu, Zhang, Biswas, and Stone]{liu2023llm+}
Bo~Liu, Yuqian Jiang, Xiaohan Zhang, Qiang Liu, Shiqi Zhang, Joydeep Biswas, and Peter Stone.
\newblock Llm+ p: Empowering large language models with optimal planning proficiency.
\newblock \emph{arXiv preprint arXiv:2304.11477}, 2023{\natexlab{a}}.

\bibitem[Liu et~al.(2023{\natexlab{b}})Liu, Yang, Shen, Hu, Zhang, Gu, and Zhang]{liu2023think}
Lei Liu, Xiaoyan Yang, Yue Shen, Binbin Hu, Zhiqiang Zhang, Jinjie Gu, and Guannan Zhang.
\newblock Think-in-memory: Recalling and post-thinking enable llms with long-term memory.
\newblock \emph{arXiv preprint arXiv:2311.08719}, 2023{\natexlab{b}}.

\bibitem[Liu et~al.(2023{\natexlab{c}})Liu, Yu, Zhang, Xu, Lei, Lai, Gu, Ding, Men, Yang, et~al.]{liu2023agentbench}
Xiao Liu, Hao Yu, Hanchen Zhang, Yifan Xu, Xuanyu Lei, Hanyu Lai, Yu~Gu, Hangliang Ding, Kaiwen Men, Kejuan Yang, et~al.
\newblock {AgentBench}: Evaluating llms as agents.
\newblock \emph{arXiv preprint arXiv:2308.03688}, 2023{\natexlab{c}}.

\bibitem[Liu et~al.(2024)Liu, Peng, Zhang, Cao, Zhang, Cheng, Wang, Yin, and Du]{liu2024tool}
Yanming Liu, Xinyue Peng, Yuwei Zhang, Jiannan Cao, Xuhong Zhang, Sheng Cheng, Xun Wang, Jianwei Yin, and Tianyu Du.
\newblock Tool-planner: Dynamic solution tree planning for large language model with tool clustering.
\newblock \emph{arXiv preprint arXiv:2406.03807}, 2024.

\bibitem[Luo et~al.(2023{\natexlab{a}})Luo, Sun, Xu, Zhao, Lou, Tao, Geng, Lin, Chen, and Zhang]{luo2023wizardmath}
Haipeng Luo, Qingfeng Sun, Can Xu, Pu~Zhao, Jianguang Lou, Chongyang Tao, Xiubo Geng, Qingwei Lin, Shifeng Chen, and Dongmei Zhang.
\newblock Wizardmath: Empowering mathematical reasoning for large language models via reinforced evol-instruct.
\newblock \emph{arXiv preprint arXiv:2308.09583}, 2023{\natexlab{a}}.

\bibitem[Luo et~al.(2023{\natexlab{b}})Luo, Xu, Zhao, Sun, Geng, Hu, Tao, Ma, Lin, and Jiang]{luo2023wizardcoder}
Ziyang Luo, Can Xu, Pu~Zhao, Qingfeng Sun, Xiubo Geng, Wenxiang Hu, Chongyang Tao, Jing Ma, Qingwei Lin, and Daxin Jiang.
\newblock Wizardcoder: Empowering code large language models with evol-instruct.
\newblock \emph{arXiv preprint arXiv:2306.08568}, 2023{\natexlab{b}}.

\bibitem[Ma et~al.(2024)Ma, Zhang, Zhu, Yang, Yang, Jin, Lan, Kong, and He]{ma2024agentboard}
Chang Ma, Junlei Zhang, Zhihao Zhu, Cheng Yang, Yujiu Yang, Yaohui Jin, Zhenzhong Lan, Lingpeng Kong, and Junxian He.
\newblock Agentboard: An analytical evaluation board of multi-turn llm agents.
\newblock \emph{arXiv preprint arXiv:2401.13178}, 2024.

\bibitem[McDermott et~al.(1998)McDermott, Ghallab, Howe, Knoblock, Ram, Veloso, Weld, and Wilkins]{McDermott1998PDDLthePD}
Drew McDermott, Malik Ghallab, Adele~E. Howe, Craig~A. Knoblock, Ashwin Ram, Manuela~M. Veloso, Daniel~S. Weld, and David~E. Wilkins.
\newblock Pddl-the planning domain definition language.
\newblock 1998.
\newblock URL \url{https://api.semanticscholar.org/CorpusID:59656859}.

\bibitem[{Meta AI}(2024)]{meta2024llama3}
{Meta AI}.
\newblock Introducing meta {Llama 3}: The most capable openly available {LLM} to date, April 2024.
\newblock URL \url{https://ai.meta.com/blog/meta-llama-3/}.
\newblock Accessed: 2024-04-18.

\bibitem[Mu et~al.(2024{\natexlab{a}})Mu, Chen, Zhang, Chen, Yu, Ge, Chen, Liang, Hu, Tao, et~al.]{mu2024robocodex}
Yao Mu, Junting Chen, Qinglong Zhang, Shoufa Chen, Qiaojun Yu, Chongjian Ge, Runjian Chen, Zhixuan Liang, Mengkang Hu, Chaofan Tao, et~al.
\newblock Robocodex: Multimodal code generation for robotic behavior synthesis.
\newblock \emph{arXiv preprint arXiv:2402.16117}, 2024{\natexlab{a}}.

\bibitem[Mu et~al.(2024{\natexlab{b}})Mu, Zhang, Hu, Wang, Ding, Jin, Wang, Dai, Qiao, and Luo]{mu2024embodiedgpt}
Yao Mu, Qinglong Zhang, Mengkang Hu, Wenhai Wang, Mingyu Ding, Jun Jin, Bin Wang, Jifeng Dai, Yu~Qiao, and Ping Luo.
\newblock Embodiedgpt: Vision-language pre-training via embodied chain of thought.
\newblock \emph{Advances in Neural Information Processing Systems}, 36, 2024{\natexlab{b}}.

\bibitem[OpenAI(2022)]{opeiai2022gpt}
OpenAI.
\newblock Openai: Introducing chatgpt, 2022.
\newblock URL \url{https://openai.com/blog/chatgpt}.

\bibitem[OpenAI(2023{\natexlab{a}})]{openai2023gpt4}
OpenAI.
\newblock Gpt-4 technical report, 2023{\natexlab{a}}.

\bibitem[OpenAI(2023{\natexlab{b}})]{openai2023gpt}
R~OpenAI.
\newblock Gpt-4 technical report. arxiv 2303.08774.
\newblock \emph{View in Article}, 2:\penalty0 13, 2023{\natexlab{b}}.

\bibitem[Parisi et~al.(2022)Parisi, Zhao, and Fiedel]{parisi2022talm}
Aaron Parisi, Yao Zhao, and Noah Fiedel.
\newblock Talm: Tool augmented language models.
\newblock \emph{arXiv preprint arXiv:2205.12255}, 2022.

\bibitem[Puig et~al.(2018)Puig, Ra, Boben, Li, Wang, Fidler, and Torralba]{puig2018virtualhome}
Xavier Puig, Kevin Ra, Marko Boben, Jiaman Li, Tingwu Wang, Sanja Fidler, and Antonio Torralba.
\newblock Virtualhome: Simulating household activities via programs.
\newblock In \emph{Proceedings of the IEEE conference on computer vision and pattern recognition}, pages 8494--8502, 2018.

\bibitem[Qin et~al.(2023)Qin, Liang, Ye, Zhu, Yan, Lu, Lin, Cong, Tang, Qian, et~al.]{qin2023toolllm}
Yujia Qin, Shihao Liang, Yining Ye, Kunlun Zhu, Lan Yan, Yaxi Lu, Yankai Lin, Xin Cong, Xiangru Tang, Bill Qian, et~al.
\newblock {ToolLLM}: Facilitating large language models to master 16000+ real-world apis.
\newblock \emph{arXiv preprint arXiv:2307.16789}, 2023.

\bibitem[Roziere et~al.(2023)Roziere, Gehring, Gloeckle, Sootla, Gat, Tan, Adi, Liu, Remez, Rapin, et~al.]{roziere2023code}
Baptiste Roziere, Jonas Gehring, Fabian Gloeckle, Sten Sootla, Itai Gat, Xiaoqing~Ellen Tan, Yossi Adi, Jingyu Liu, Tal Remez, J{\'e}r{\'e}my Rapin, et~al.
\newblock Code llama: Open foundation models for code.
\newblock \emph{arXiv preprint arXiv:2308.12950}, 2023.

\bibitem[Ruan et~al.(2023)Ruan, Chen, Zhang, Xu, Bao, Du, Shi, Mao, Zeng, and Zhao]{ruan2023tptu}
Jingqing Ruan, Yihong Chen, Bin Zhang, Zhiwei Xu, Tianpeng Bao, Guoqing Du, Shiwei Shi, Hangyu Mao, Xingyu Zeng, and Rui Zhao.
\newblock Tptu: Task planning and tool usage of large language model-based ai agents.
\newblock \emph{arXiv preprint arXiv:2308.03427}, 2023.

\bibitem[Russell and Norvig(2016)]{russell2016artificial}
Stuart~J Russell and Peter Norvig.
\newblock \emph{Artificial intelligence: a modern approach}.
\newblock Pearson, 2016.

\bibitem[Schick et~al.(2023)Schick, Dwivedi{-}Yu, Dess{\`{\i}}, Raileanu, Lomeli, Zettlemoyer, Cancedda, and Scialom]{TimoSchick2023_Toolformer}
Timo Schick, Jane Dwivedi{-}Yu, Roberto Dess{\`{\i}}, Roberta Raileanu, Maria Lomeli, Luke Zettlemoyer, Nicola Cancedda, and Thomas Scialom.
\newblock Toolformer: Language models can teach themselves to use tools.
\newblock \emph{CoRR}, abs/2302.04761, 2023.
\newblock \doi{10.48550/ARXIV.2302.04761}.
\newblock URL \url{https://doi.org/10.48550/arXiv.2302.04761}.

\bibitem[Shen et~al.(2023)Shen, Song, Tan, Li, Lu, and Zhuang]{YongliangShen2023_HuggingGPT}
Yongliang Shen, Kaitao Song, Xu~Tan, Dongsheng Li, Weiming Lu, and Yueting Zhuang.
\newblock Hugginggpt: Solving {AI} tasks with chatgpt and its friends in huggingface.
\newblock \emph{CoRR}, abs/2303.17580, 2023.
\newblock \doi{10.48550/ARXIV.2303.17580}.
\newblock URL \url{https://doi.org/10.48550/arXiv.2303.17580}.

\bibitem[Shinn et~al.(2023)Shinn, Cassano, Gopinath, Narasimhan, and Yao]{shinn2023reflexion}
Noah Shinn, Federico Cassano, Ashwin Gopinath, Karthik~R Narasimhan, and Shunyu Yao.
\newblock Reflexion: Language agents with verbal reinforcement learning.
\newblock In \emph{Thirty-seventh Conference on Neural Information Processing Systems}, 2023.

\bibitem[Shridhar et~al.(2020)Shridhar, Yuan, C{\^o}t{\'e}, Bisk, Trischler, and Hausknecht]{shridhar2020alfworld}
Mohit Shridhar, Xingdi Yuan, Marc-Alexandre C{\^o}t{\'e}, Yonatan Bisk, Adam Trischler, and Matthew Hausknecht.
\newblock Alfworld: Aligning text and embodied environments for interactive learning.
\newblock \emph{arXiv preprint arXiv:2010.03768}, 2020.

\bibitem[Singh et~al.(2023)Singh, Blukis, Mousavian, Goyal, Xu, Tremblay, Fox, Thomason, and Garg]{singh2023progprompt}
Ishika Singh, Valts Blukis, Arsalan Mousavian, Ankit Goyal, Danfei Xu, Jonathan Tremblay, Dieter Fox, Jesse Thomason, and Animesh Garg.
\newblock Progprompt: Generating situated robot task plans using large language models.
\newblock In \emph{2023 IEEE International Conference on Robotics and Automation (ICRA)}, pages 11523--11530. IEEE, 2023.

\bibitem[Song et~al.(2023)Song, Wu, Washington, Sadler, Chao, and Su]{song2023llmplanner}
Chan~Hee Song, Jiaman Wu, Clayton Washington, Brian~M. Sadler, Wei-Lun Chao, and Yu~Su.
\newblock Llm-planner: Few-shot grounded planning for embodied agents with large language models, 2023.

\bibitem[Song et~al.(2024)Song, Yin, Yue, Huang, Li, and Lin]{song2024trial}
Yifan Song, Da~Yin, Xiang Yue, Jie Huang, Sujian Li, and Bill~Yuchen Lin.
\newblock Trial and error: Exploration-based trajectory optimization for llm agents.
\newblock \emph{arXiv preprint arXiv:2403.02502}, 2024.

\bibitem[Sumers et~al.(2023)Sumers, Yao, Narasimhan, and Griffiths]{sumers2023cognitive}
Theodore~R Sumers, Shunyu Yao, Karthik Narasimhan, and Thomas~L Griffiths.
\newblock Cognitive architectures for language agents.
\newblock \emph{arXiv preprint arXiv:2309.02427}, 2023.

\bibitem[Sun et~al.(2023{\natexlab{a}})Sun, Zhuang, Kong, Dai, and Zhang]{sun2023adaplanner}
Haotian Sun, Yuchen Zhuang, Lingkai Kong, Bo~Dai, and Chao Zhang.
\newblock Adaplanner: Adaptive planning from feedback with language models.
\newblock \emph{arXiv preprint arXiv:2305.16653}, 2023{\natexlab{a}}.

\bibitem[Sun et~al.(2023{\natexlab{b}})Sun, Liu, Wang, Zhu, and Iyyer]{sun2023pearl}
Simeng Sun, Yang Liu, Shuohang Wang, Chenguang Zhu, and Mohit Iyyer.
\newblock Pearl: Prompting large language models to plan and execute actions over long documents.
\newblock \emph{arXiv preprint arXiv:2305.14564}, 2023{\natexlab{b}}.

\bibitem[Tack et~al.(2024)Tack, Kim, Mitchell, Shin, Teh, and Schwarz]{tack2024online}
Jihoon Tack, Jaehyung Kim, Eric Mitchell, Jinwoo Shin, Yee~Whye Teh, and Jonathan~Richard Schwarz.
\newblock Online adaptation of language models with a memory of amortized contexts.
\newblock \emph{arXiv preprint arXiv:2403.04317}, 2024.

\bibitem[Touvron et~al.(2023)Touvron, Martin, Stone, Albert, Almahairi, Babaei, Bashlykov, Batra, Bhargava, Bhosale, et~al.]{touvron2023llama}
Hugo Touvron, Louis Martin, Kevin Stone, Peter Albert, Amjad Almahairi, Yasmine Babaei, Nikolay Bashlykov, Soumya Batra, Prajjwal Bhargava, Shruti Bhosale, et~al.
\newblock Llama 2: Open foundation and fine-tuned chat models.
\newblock \emph{arXiv preprint arXiv:2307.09288}, 2023.

\bibitem[Valmeekam et~al.(2024)Valmeekam, Marquez, Olmo, Sreedharan, and Kambhampati]{valmeekam2024planbench}
Karthik Valmeekam, Matthew Marquez, Alberto Olmo, Sarath Sreedharan, and Subbarao Kambhampati.
\newblock Planbench: An extensible benchmark for evaluating large language models on planning and reasoning about change.
\newblock \emph{Advances in Neural Information Processing Systems}, 36, 2024.

\bibitem[Wang et~al.(2024{\natexlab{a}})Wang, Fang, Eisner, Van~Durme, and Su]{wang2024llms}
Boshi Wang, Hao Fang, Jason Eisner, Benjamin Van~Durme, and Yu~Su.
\newblock Llms in the imaginarium: tool learning through simulated trial and error.
\newblock \emph{arXiv preprint arXiv:2403.04746}, 2024{\natexlab{a}}.

\bibitem[Wang et~al.(2023{\natexlab{a}})Wang, Ma, Feng, Zhang, Yang, Zhang, Chen, Tang, Chen, Lin, et~al.]{wang2023survey}
Lei Wang, Chen Ma, Xueyang Feng, Zeyu Zhang, Hao Yang, Jingsen Zhang, Zhiyuan Chen, Jiakai Tang, Xu~Chen, Yankai Lin, et~al.
\newblock A survey on large language model based autonomous agents.
\newblock \emph{arXiv preprint arXiv:2308.11432}, 2023{\natexlab{a}}.

\bibitem[Wang et~al.(2023{\natexlab{b}})Wang, Xu, Lan, Hu, Lan, Lee, and Lim]{wang2023plan}
Lei Wang, Wanyu Xu, Yihuai Lan, Zhiqiang Hu, Yunshi Lan, Roy Ka-Wei Lee, and Ee-Peng Lim.
\newblock Plan-and-solve prompting: Improving zero-shot chain-of-thought reasoning by large language models.
\newblock \emph{arXiv preprint arXiv:2305.04091}, 2023{\natexlab{b}}.

\bibitem[Wang et~al.(2023{\natexlab{c}})Wang, Ling, Yuan, Shridhar, Bao, Qin, Wang, Xu, and Wang]{wang2023gensim}
Lirui Wang, Yiyang Ling, Zhecheng Yuan, Mohit Shridhar, Chen Bao, Yuzhe Qin, Bailin Wang, Huazhe Xu, and Xiaolong Wang.
\newblock Gensim: Generating robotic simulation tasks via large language models.
\newblock \emph{arXiv preprint arXiv:2310.01361}, 2023{\natexlab{c}}.

\bibitem[Wang et~al.(2024{\natexlab{b}})Wang, Li, Han, Zhang, and Baldwin]{wang2024learning}
Renxi Wang, Haonan Li, Xudong Han, Yixuan Zhang, and Timothy Baldwin.
\newblock Learning from failure: Integrating negative examples when fine-tuning large language models as agents.
\newblock \emph{arXiv preprint arXiv:2402.11651}, 2024{\natexlab{b}}.

\bibitem[Wang et~al.(2023{\natexlab{d}})Wang, Todd, Yuan, Xiao, C{\^o}t{\'e}, and Jansen]{wang2023bytesized32}
Ruoyao Wang, Graham Todd, Eric Yuan, Ziang Xiao, Marc-Alexandre C{\^o}t{\'e}, and Peter Jansen.
\newblock Bytesized32: A corpus and challenge task for generating task-specific world models expressed as text games.
\newblock \emph{arXiv preprint arXiv:2305.14879}, 2023{\natexlab{d}}.

\bibitem[Wang et~al.(2023{\natexlab{e}})Wang, Li, Wang, Bai, Luo, Zhang, Jojic, Xing, and Hu]{wang2023promptagent}
Xinyuan Wang, Chenxi Li, Zhen Wang, Fan Bai, Haotian Luo, Jiayou Zhang, Nebojsa Jojic, Eric~P Xing, and Zhiting Hu.
\newblock Promptagent: Strategic planning with language models enables expert-level prompt optimization.
\newblock \emph{arXiv preprint arXiv:2310.16427}, 2023{\natexlab{e}}.

\bibitem[Wang et~al.(2023{\natexlab{f}})Wang, Xian, Chen, Wang, Wang, Fragkiadaki, Erickson, Held, and Gan]{wang2023robogen}
Yufei Wang, Zhou Xian, Feng Chen, Tsun-Hsuan Wang, Yian Wang, Katerina Fragkiadaki, Zackory Erickson, David Held, and Chuang Gan.
\newblock Robogen: Towards unleashing infinite data for automated robot learning via generative simulation.
\newblock \emph{arXiv preprint arXiv:2311.01455}, 2023{\natexlab{f}}.

\bibitem[Wang et~al.(2023{\natexlab{g}})Wang, Cai, Liu, Ma, and Liang]{wang2023describe}
Zihao Wang, Shaofei Cai, Anji Liu, Xiaojian Ma, and Yitao Liang.
\newblock Describe, explain, plan and select: Interactive planning with large language models enables open-world multi-task agents, 2023{\natexlab{g}}.

\bibitem[Wei et~al.(2021)Wei, Bosma, Zhao, Guu, Yu, Lester, Du, Dai, and Le]{wei2021finetuned}
Jason Wei, Maarten Bosma, Vincent~Y Zhao, Kelvin Guu, Adams~Wei Yu, Brian Lester, Nan Du, Andrew~M Dai, and Quoc~V Le.
\newblock Finetuned language models are zero-shot learners.
\newblock \emph{arXiv preprint arXiv:2109.01652}, 2021.

\bibitem[Wei et~al.(2022)Wei, Wang, Schuurmans, Bosma, Xia, Chi, Le, Zhou, et~al.]{wei2022chain}
Jason Wei, Xuezhi Wang, Dale Schuurmans, Maarten Bosma, Fei Xia, Ed~Chi, Quoc~V Le, Denny Zhou, et~al.
\newblock Chain-of-thought prompting elicits reasoning in large language models.
\newblock \emph{Advances in Neural Information Processing Systems}, 35:\penalty0 24824--24837, 2022.

\bibitem[Wu et~al.(2023{\natexlab{a}})Wu, Bansal, Zhang, Wu, Zhang, Zhu, Li, Jiang, Zhang, and Wang]{wu2023autogen}
Qingyun Wu, Gagan Bansal, Jieyu Zhang, Yiran Wu, Shaokun Zhang, Erkang Zhu, Beibin Li, Li~Jiang, Xiaoyun Zhang, and Chi Wang.
\newblock {AutoGen}: Enabling next-gen llm applications via multi-agent conversation framework.
\newblock \emph{arXiv preprint arXiv:2308.08155}, 2023{\natexlab{a}}.

\bibitem[Wu et~al.(2023{\natexlab{b}})Wu, Wang, Xu, Lu, and Yan]{wu2023embodied}
Zhenyu Wu, Ziwei Wang, Xiuwei Xu, Jiwen Lu, and Haibin Yan.
\newblock Embodied task planning with large language models, 2023{\natexlab{b}}.

\bibitem[Xi et~al.(2023)Xi, Chen, Guo, He, Ding, Hong, Zhang, Wang, Jin, Zhou, et~al.]{xi2023rise}
Zhiheng Xi, Wenxiang Chen, Xin Guo, Wei He, Yiwen Ding, Boyang Hong, Ming Zhang, Junzhe Wang, Senjie Jin, Enyu Zhou, et~al.
\newblock The rise and potential of large language model based agents: A survey.
\newblock \emph{arXiv preprint arXiv:2309.07864}, 2023.

\bibitem[Xie et~al.(2024)Xie, Zhang, Chen, Zhu, Lou, Tian, Xiao, and Su]{xie2024travelplanner}
Jian Xie, Kai Zhang, Jiangjie Chen, Tinghui Zhu, Renze Lou, Yuandong Tian, Yanghua Xiao, and Yu~Su.
\newblock Travelplanner: A benchmark for real-world planning with language agents.
\newblock \emph{arXiv preprint arXiv:2402.01622}, 2024.

\bibitem[Xie et~al.(2023)Xie, Zhou, Cheng, Shi, Weng, Liu, Hua, Zhao, Liu, Liu, Liu, Xu, Su, Shin, Xiong, and Yu]{TianbaoXie2023_OpenAgents}
Tianbao Xie, Fan Zhou, Zhoujun Cheng, Peng Shi, Luoxuan Weng, Yitao Liu, Toh~Jing Hua, Junning Zhao, Qian Liu, Che Liu, Leo~Z. Liu, Yiheng Xu, Hongjin Su, Dongchan Shin, Caiming Xiong, and Tao Yu.
\newblock Openagents: An open platform for language agents in the wild.
\newblock \emph{CoRR}, abs/2310.10634, 2023.
\newblock \doi{10.48550/ARXIV.2310.10634}.
\newblock URL \url{https://doi.org/10.48550/arXiv.2310.10634}.

\bibitem[Xu et~al.(2023)Xu, Sun, Zheng, Geng, Zhao, Feng, Tao, and Jiang]{xu2023wizardlm}
Can Xu, Qingfeng Sun, Kai Zheng, Xiubo Geng, Pu~Zhao, Jiazhan Feng, Chongyang Tao, and Daxin Jiang.
\newblock Wizardlm: Empowering large language models to follow complex instructions.
\newblock \emph{arXiv preprint arXiv:2304.12244}, 2023.

\bibitem[Yang et~al.(2024)Yang, Sun, Weihs, VanderBilt, Herrasti, Han, Wu, Haber, Krishna, Liu, et~al.]{yang2024holodeck}
Yue Yang, Fan-Yun Sun, Luca Weihs, Eli VanderBilt, Alvaro Herrasti, Winson Han, Jiajun Wu, Nick Haber, Ranjay Krishna, Lingjie Liu, et~al.
\newblock Holodeck: Language guided generation of 3d embodied ai environments.
\newblock In \emph{Proceedings of the IEEE/CVF Conference on Computer Vision and Pattern Recognition}, pages 16227--16237, 2024.

\bibitem[Yao et~al.(2022)Yao, Zhao, Yu, Du, Shafran, Narasimhan, and Cao]{yao2022react}
Shunyu Yao, Jeffrey Zhao, Dian Yu, Nan Du, Izhak Shafran, Karthik Narasimhan, and Yuan Cao.
\newblock React: Synergizing reasoning and acting in language models.
\newblock \emph{arXiv preprint arXiv:2210.03629}, 2022.

\bibitem[Yao et~al.(2023{\natexlab{a}})Yao, Yu, Zhao, Shafran, Griffiths, Cao, and Narasimhan]{yao2023tree}
Shunyu Yao, Dian Yu, Jeffrey Zhao, Izhak Shafran, Thomas~L Griffiths, Yuan Cao, and Karthik Narasimhan.
\newblock Tree of thoughts: Deliberate problem solving with large language models.
\newblock \emph{arXiv preprint arXiv:2305.10601}, 2023{\natexlab{a}}.

\bibitem[Yao et~al.(2023{\natexlab{b}})Yao, Heinecke, Niebles, Liu, Feng, Xue, Murthy, Chen, Zhang, Arpit, et~al.]{yao2023retroformer}
Weiran Yao, Shelby Heinecke, Juan~Carlos Niebles, Zhiwei Liu, Yihao Feng, Le~Xue, Rithesh Murthy, Zeyuan Chen, Jianguo Zhang, Devansh Arpit, et~al.
\newblock Retroformer: Retrospective large language agents with policy gradient optimization.
\newblock \emph{arXiv preprint arXiv:2308.02151}, 2023{\natexlab{b}}.

\bibitem[Yin et~al.(2023)Yin, Brahman, Ravichander, Chandu, Chang, Choi, and Lin]{yin2023lumos}
Da~Yin, Faeze Brahman, Abhilasha Ravichander, Khyathi Chandu, Kai-Wei Chang, Yejin Choi, and Bill~Yuchen Lin.
\newblock Lumos: Learning agents with unified data, modular design, and open-source llms.
\newblock \emph{arXiv preprint arXiv:2311.05657}, 2023.

\bibitem[Zeng et~al.(2023)Zeng, Liu, Lu, Wang, Liu, Dong, and Tang]{zeng2023agenttuning}
Aohan Zeng, Mingdao Liu, Rui Lu, Bowen Wang, Xiao Liu, Yuxiao Dong, and Jie Tang.
\newblock Agenttuning: Enabling generalized agent abilities for llms.
\newblock \emph{arXiv preprint arXiv:2310.12823}, 2023.

\bibitem[Zhang et~al.(2024)Zhang, Lan, Murthy, Liu, Yao, Tan, Hoang, Yang, Feng, Liu, et~al.]{zhang2024agentohana}
Jianguo Zhang, Tian Lan, Rithesh Murthy, Zhiwei Liu, Weiran Yao, Juntao Tan, Thai Hoang, Liangwei Yang, Yihao Feng, Zuxin Liu, et~al.
\newblock Agentohana: Design unified data and training pipeline for effective agent learning.
\newblock \emph{arXiv preprint arXiv:2402.15506}, 2024.

\bibitem[Zhao et~al.(2024{\natexlab{a}})Zhao, Huang, Xu, Lin, Liu, and Huang]{zhao2024expel}
Andrew Zhao, Daniel Huang, Quentin Xu, Matthieu Lin, Yong-Jin Liu, and Gao Huang.
\newblock Expel: Llm agents are experiential learners.
\newblock In \emph{Proceedings of the AAAI Conference on Artificial Intelligence}, volume~38, pages 19632--19642, 2024{\natexlab{a}}.

\bibitem[Zhao et~al.(2024{\natexlab{b}})Zhao, Lee, and Hsu]{zhao2024large}
Zirui Zhao, Wee~Sun Lee, and David Hsu.
\newblock Large language models as commonsense knowledge for large-scale task planning.
\newblock \emph{Advances in Neural Information Processing Systems}, 36, 2024{\natexlab{b}}.

\bibitem[Zheng et~al.(2024)Zheng, Mishra, Zhang, Chen, Chen, Nova, Hou, Cheng, Le, Chi, et~al.]{zheng2024natural}
Huaixiu~Steven Zheng, Swaroop Mishra, Hugh Zhang, Xinyun Chen, Minmin Chen, Azade Nova, Le~Hou, Heng-Tze Cheng, Quoc~V Le, Ed~H Chi, et~al.
\newblock Natural plan: Benchmarking llms on natural language planning.
\newblock \emph{arXiv preprint arXiv:2406.04520}, 2024.

\bibitem[Zhong et~al.(2024)Zhong, Guo, Gao, Ye, and Wang]{zhong2024memorybank}
Wanjun Zhong, Lianghong Guo, Qiqi Gao, He~Ye, and Yanlin Wang.
\newblock Memorybank: Enhancing large language models with long-term memory.
\newblock In \emph{Proceedings of the AAAI Conference on Artificial Intelligence}, volume~38, pages 19724--19731, 2024.

\bibitem[Zhou et~al.(2023)Zhou, Yan, Shlapentokh-Rothman, Wang, and Wang]{zhou2023language}
Andy Zhou, Kai Yan, Michal Shlapentokh-Rothman, Haohan Wang, and Yu-Xiong Wang.
\newblock Language agent tree search unifies reasoning acting and planning in language models.
\newblock \emph{arXiv preprint arXiv:2310.04406}, 2023.

\bibitem[Zhou et~al.(2024)Zhou, Liu, Xu, Iyer, Sun, Mao, Ma, Efrat, Yu, Yu, et~al.]{zhou2024lima}
Chunting Zhou, Pengfei Liu, Puxin Xu, Srinivasan Iyer, Jiao Sun, Yuning Mao, Xuezhe Ma, Avia Efrat, Ping Yu, Lili Yu, et~al.
\newblock Lima: Less is more for alignment.
\newblock \emph{Advances in Neural Information Processing Systems}, 36, 2024.

\bibitem[Zhou et~al.(2022)Zhou, Sch{\"a}rli, Hou, Wei, Scales, Wang, Schuurmans, Cui, Bousquet, Le, et~al.]{zhou2022least}
Denny Zhou, Nathanael Sch{\"a}rli, Le~Hou, Jason Wei, Nathan Scales, Xuezhi Wang, Dale Schuurmans, Claire Cui, Olivier Bousquet, Quoc Le, et~al.
\newblock Least-to-most prompting enables complex reasoning in large language models.
\newblock \emph{arXiv preprint arXiv:2205.10625}, 2022.

\bibitem[Zhu et~al.(2023)Zhu, Chen, Tian, Tao, Su, Yang, Huang, Li, Lu, Wang, et~al.]{zhu2023ghost}
Xizhou Zhu, Yuntao Chen, Hao Tian, Chenxin Tao, Weijie Su, Chenyu Yang, Gao Huang, Bin Li, Lewei Lu, Xiaogang Wang, et~al.
\newblock Ghost in the minecraft: Generally capable agents for open-world environments via large language models with text-based knowledge and memory.
\newblock \emph{arXiv preprint arXiv:2305.17144}, 2023.

\end{thebibliography}

\newpage
\appendix
\tcbset{colframe = blue!50!black, 
colback = white, 
toptitle=1mm, 
bottomtitle=1mm, 
fonttitle=\fontsize{10pt}{12pt}\selectfont,  breakable, enhanced jigsaw, finish broken={\tcbset{colframe=white, bottom=0mm, bottomrule=0mm}},before upper={\tcbset{colframe=white, toprule=0mm}},after={\tcbset{colframe=blue!40!black, toprule=1mm, bottomrule=1mm}},}

\section{More Implementation Details}

\subsection{Models}
\label{app:models}

We applied the instruct version for models. Specifically, the detailed version for each model is presented in Table~\ref{tab:models}.

\begin{table}[h]
\centering
\begin{tabular}{ll} 
\toprule
Model & Version  \\
\midrule
CodeLlama & \href{https://huggingface.co/meta-llama/CodeLlama-7b-Instruct-hf}{meta-llama/CodeLlama-7b-Instruct-hf} \\
Mistral & \href{https://huggingface.co/mistralai/Mistral-7B-Instruct-v0.2}{mistralai/Mistral-7B-Instruct-v0.2} \\
Llama2 & \href{https://huggingface.co/meta-llama/Llama-2-7b-chat-hf}{meta-llama/Llama-2-7b-chat-hf} \\
Llama3 & \href{https://huggingface.co/meta-llama/Meta-Llama-3-8B-Instruct}{meta-llama/Meta-Llama-3-8B-Instruct} \\
AgentLM & \href{https://huggingface.co/THUDM/agentlm-7b}{THUDM/agentlm-7b} \\
\bottomrule
\end{tabular}
\caption{Evaluated models in this study.}
\label{tab:models}
\end{table}

\subsection{Natural Language Mapping}
\label{app:nl_map}

We leverage GPT-4 to generate the natural language mapping that converts structured actions into its natural language format. When the mapping failed to yield, we heuristically serialized the structured actions.
The prompt for generating natural language mapping with GPT-4 is as follows:

\begin{tcolorbox}[colback=gray!10!white,colframe=gray!50!white,title=Natural Language Mapping Generation]
I would like you to create natural language mapping for PDDL.   \\
The form of the natural language mapping is a Python dictionary, wherein\\
1. The key corresponds to the name of a predicate or action within the domain PDDL.\\
2. The value is its equivalent in natural language, with parameters presented in "\{argn\}", where n is the index of its parameters in the PDDL expression.\\
3. You must ensure that the number of "\{\}" corresponds precisely to the number of parameters in predicates or actions.  \\
4. You should very carefully check the order of \{argn\}. \\
\\
Your output must strictly follow the provided example.\\
\\
Example:\\
PDDL Domain:\\
```pddl\\
(define (domain hanoi)\\
  (:requirements :strips)\\
  (:predicates\\
  (clear ?x)\\
  (on ?x ?y)\\
  (smaller ?x ?y)\\
  )\\
\\
  (:action move\\
    :parameters (?disc ?from ?to)\\
    :precondition (and (smaller ?to ?disc) (on ?disc ?from)\\
               (clear ?disc) (clear ?to))\\
    :effect  (and (clear ?from) (on ?disc ?to) (not (on ?disc ?from))\\
          (not (clear ?to))))\\
  )\\
```\\
Specification:\\
Your goal is to solve the Tower of Hanoi puzzle, which involves moving a stack of discs from one peg to another, with the restriction that no disc may be placed on top of a smaller disc. The puzzle is solved when all the discs are moved to the target peg following these rules.\\
\\
The actions defined in this domain include:\\
- move <disc> <from> <to>: This action allows moving a disc from one peg to another. The preconditions for this action are that the target peg is smaller than the disc being moved, the disc is on the source peg, and both the disc and the target peg are clear (i.e., there is no disc on top of them). The effect of this action is that the source peg becomes clear, the disc is now on the target peg, the disc is no longer on the source peg, and the target peg is no longer clear.\\
\\
You have the following restrictions on your actions:\\
- A disc can only be moved if it is clear, meaning there is no other disc on top of it.\\
- A disc can only be placed on another disc or peg that is larger than itself.\\
- A disc can only be moved to a peg that is clear.\\
- Once a disc is moved from a peg, that peg becomes clear.\\
- Once a disc is placed on a peg, that peg is no longer clear.\\
 Natural Language Mapping:\\
```python\\
\{\\
    "clear": "\{arg1\} is clear.",\\
    "on": "\{arg1\} is on \{arg2\}.",\\
    "smaller": "\{arg1\} is smaller than \{arg2\}.",\\
    "move": "Move \{arg1\} from \{arg2\} to \{arg3\}."\\
\}\\
```\\
\\
You need to generate the corresponding natural language mapping for the following pddl domain.\\
    \\
PDDL Domain:\\
\{PDDL\_Domain\}\\
Specification:\\
\{PDDL\_Description\}\\
Natural Language Mapping:\\
\end{tcolorbox}

\section{More Statistics on Environment}
\label{app:stat}

\subsection{Environment Specification}

\begin{figure}[htbp]
    \centering
    \includegraphics[width=\linewidth]{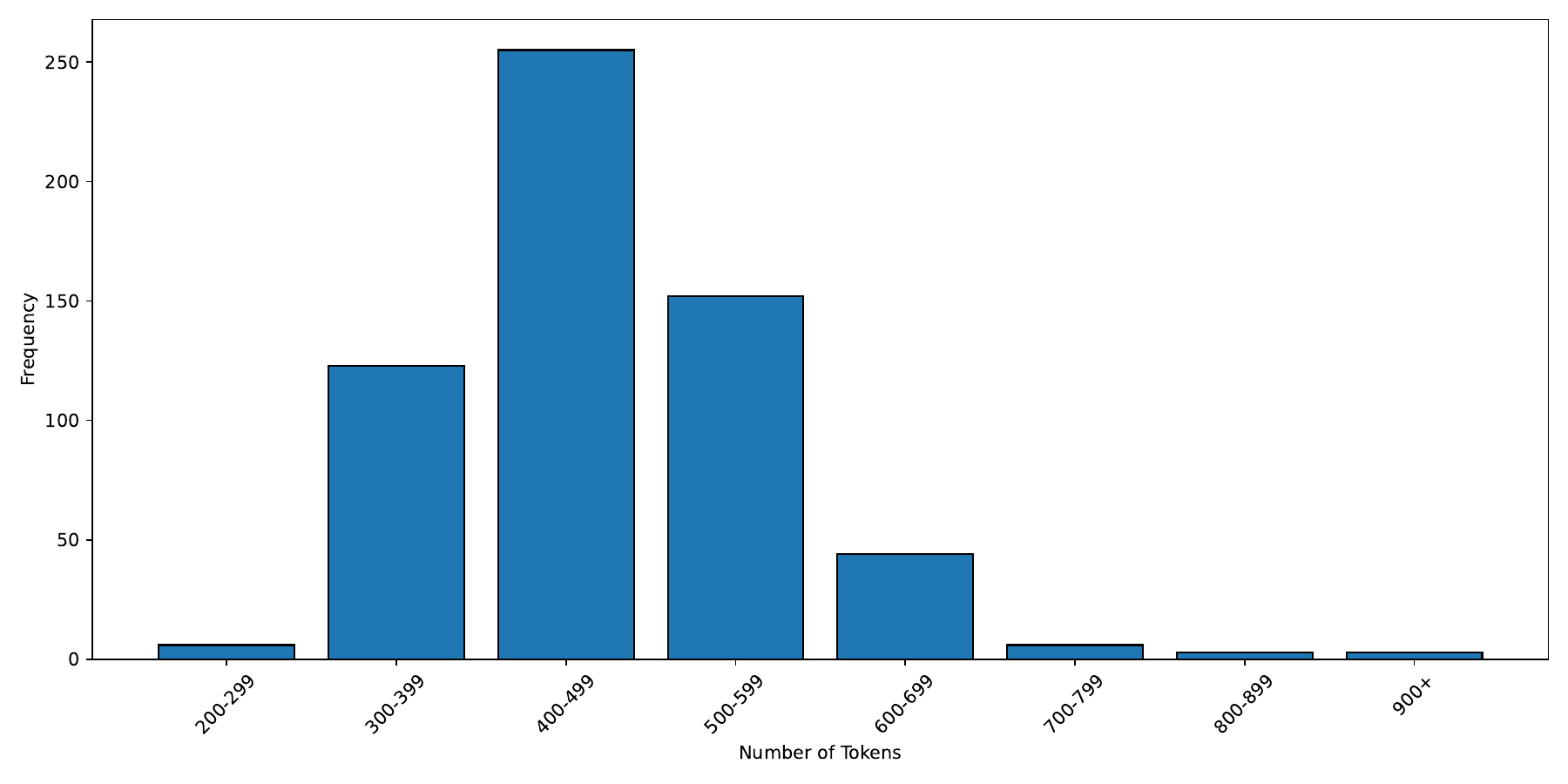}
    \caption{
    The token distribution of the generated environment specification.
    }
    \label{fig:token_distribution_specification}
\end{figure}

We analyzed the token distribution within the environmental specifications. Among the 592 environmental specifications, the average token count is 473.55, with a median of 467.00. The minimum token count is 207, and the maximum is 934. As depicted in Figure~\ref{fig:token_distribution_specification}, the number of specification tokens for the environment is predominantly concentrated within the range of 300 to 699.

\subsection{Environment Implementation}



\begin{figure}[htbp]
    \centering
    \includegraphics[width=\linewidth]{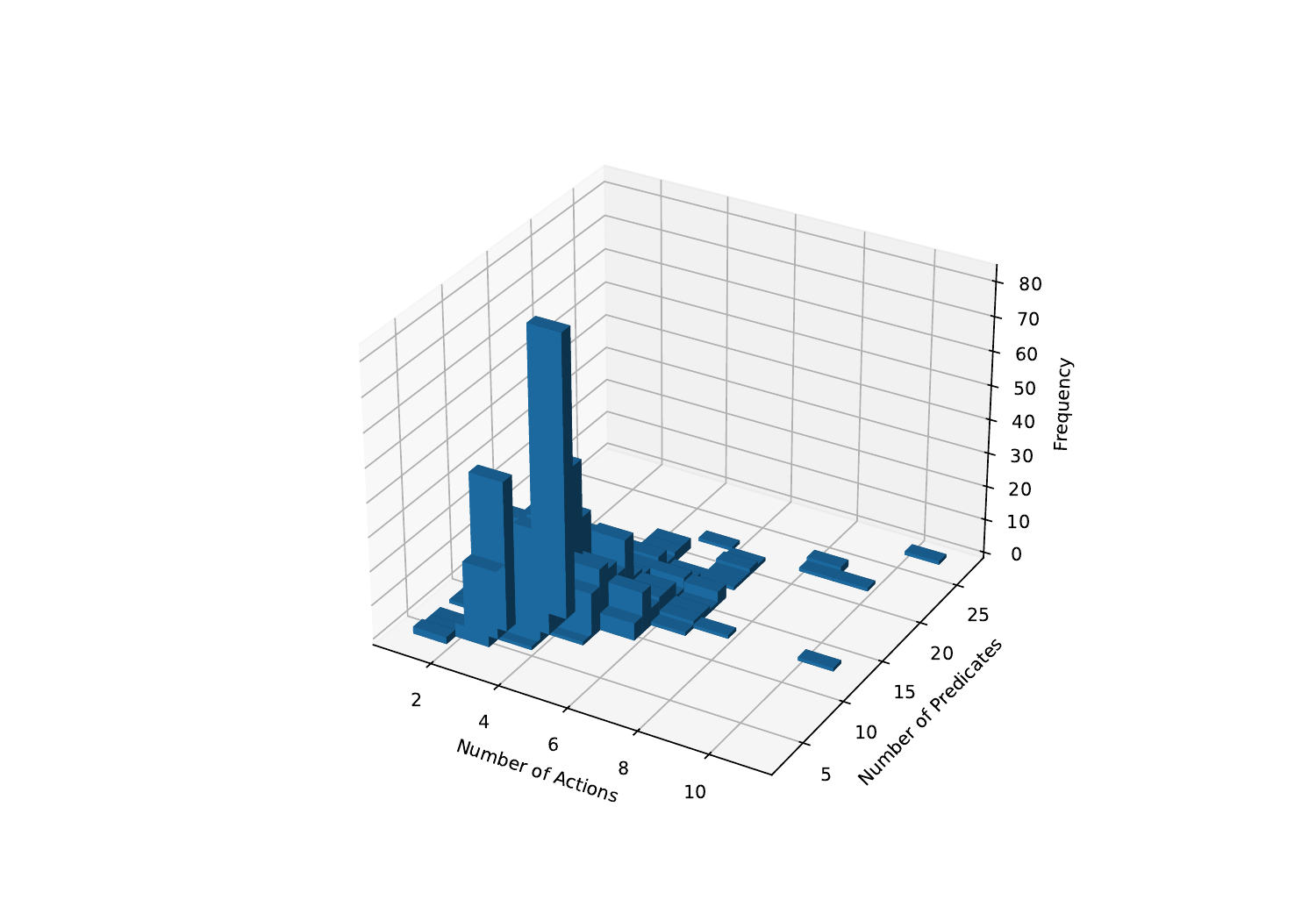}
    \caption{
    The frequency distribution of actions and predicates in datasets.
    }
    \label{fig:frequency_distribution_3d}
\end{figure}

The scale of action space and state space in an environment typically dictates its complexity, with a greater number of actions and states generally indicating a more complex environment. 
An environment library with a greater variety of difficulty levels is preferable for a training set. 
As shown in Figure~\ref{fig:frequency_distribution_3d}, there is a significant diversity in the number of actions and predicates.

\subsection{Diversity Analysis}

\begin{figure}[htbp]
    \centering
    \includegraphics[width=\linewidth]{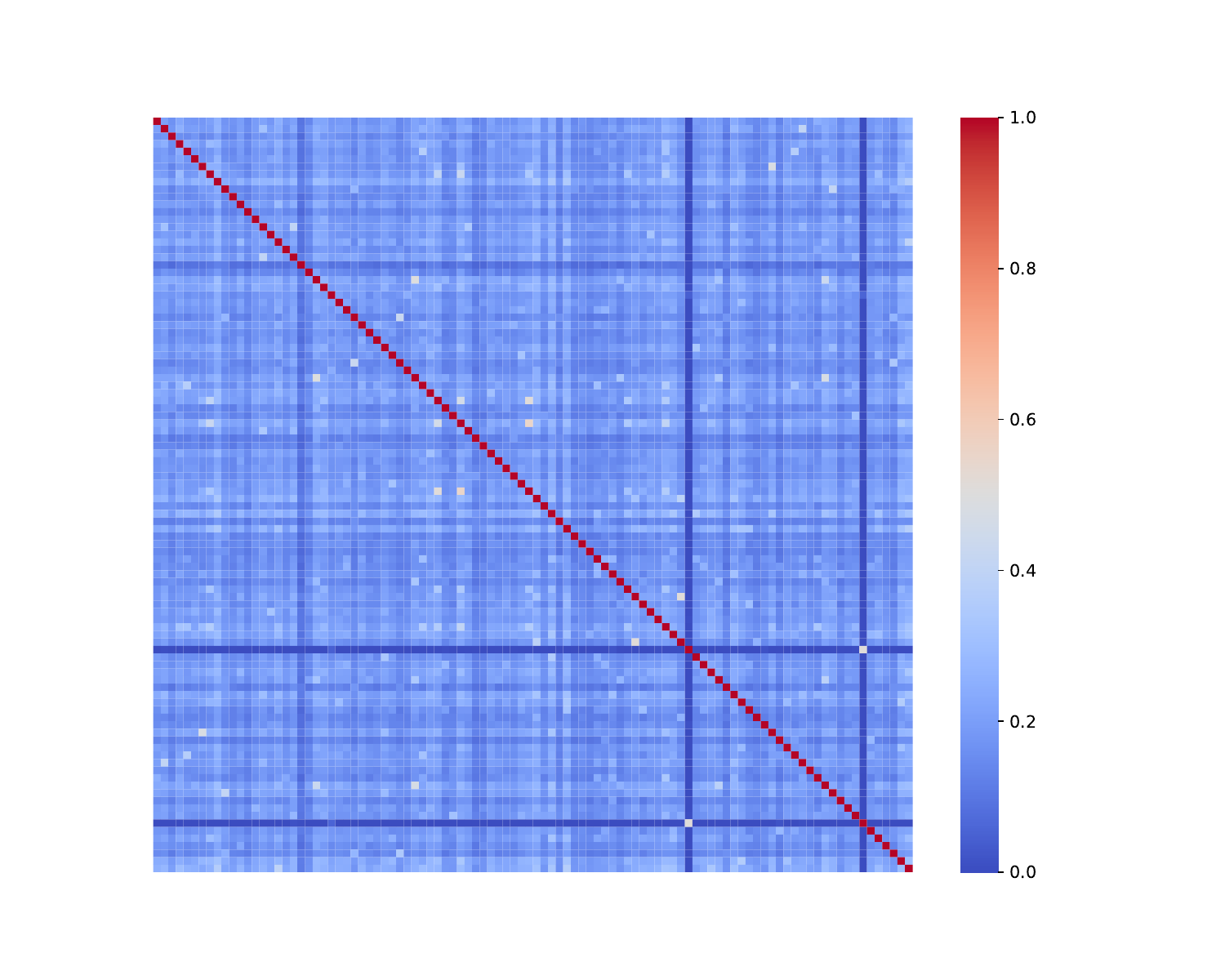}
    \caption{
    Cosine similarity heatmap depicting the semantic relationships among randomly sampled 100 environment specifications. 
    Darker shades represent a higher similarity between the two specifications. 
    }
    \label{fig:heatmap}
\end{figure}
We evaluate the diversity of generated environments using cosine similarity. More specifically, we randomly sampled 100 environment specifications for better visualization and converted them into TF-IDF vectors. After calculating the cosine similarity matrix between all pairs of specifications, we visualize the matrix using heatmap as is shown in Figure~\ref{fig:heatmap}.
The computed average cosine similarity of the sampled environment specifications is 0.176, indicating that the corpus exhibits a high degree of diversity, reflecting a rich tapestry of distinct semantic features and thematic elements.

\section{Examples}
\label{app:examples}

In this section, we present the specific details of the cases depicted in Figure~\ref{fig:environment_generation} and Figure~\ref{fig:goal_generation}.

\subsection{Environment Specification}

\begin{tcolorbox}[colback=gray!10!white,colframe=gray!50!white,title=Environment Specification]
You are a nutritionist tasked with creating a new healthy recipe book that incorporates peanut butter powder as a key ingredient. Your environment consists of your office, a computer for research and writing, a kitchen for testing recipes, and a collection of ingredients including peanut butter powder. The computer can be used to write and save recipes, but it needs to be charged after every three hours of use.\\
\\
The actions defined in this domain include:\\
- \*\*research\_ingredient <location> <ingredient> <nutritionist>\*\*: This action allows the nutritionist to research peanut butter powder at the office. The effect of this action is that the nutritionist gains knowledge about the nutritional benefits and potential uses of peanut butter powder in recipes.\\
- \*\*develop\_recipe <location> <computer> <nutritionist> <ingredient>\*\*: This action allows the nutritionist to create a new recipe using peanut butter powder. The precondition for this action is that the nutritionist has researched peanut butter powder. The effect of this action is a draft of a new recipe that incorporates peanut butter powder.\\
- \*\*test\_recipe <location> <kitchen> <nutritionist> <recipe>\*\*: This action allows the nutritionist to test the drafted recipe in the kitchen. The preconditions for this action are that the nutritionist has developed a recipe and has all the necessary ingredients, including peanut butter powder. The effect of this action is feedback on the recipe's taste, nutritional value, and overall feasibility.\\
- \*\*finalize\_recipe <location> <computer> <nutritionist> <tested\_recipe>\*\*: This action allows the nutritionist to finalize the recipe after testing and receiving feedback. The preconditions for this action are that the nutritionist has tested the recipe and the computer is charged. The effect of this action is a completed recipe that is ready to be included in the healthy recipe book.\\
\\
You have the following restrictions on your actions:\\
- You can only research peanut butter powder if you are in the office where your research materials are located.\\
- You can only develop a recipe after researching peanut butter powder and must have a charged computer.\\
- You can only test a recipe in the kitchen if you have developed a recipe and have all the necessary ingredients.\\
- You can only finalize a recipe after testing it in the kitchen and receiving feedback, and if your computer is charged.\\
\end{tcolorbox}

\subsection{Environment Implementation}

\begin{tcolorbox}[colback=gray!10!white,colframe=gray!50!white,title=Environment Implementation, sharp corners=south]
\begin{minipage}[t]{.95\textwidth}
\small 
\begin{Verbatim}[commandchars=\\\{\}]
(define (domain healthy-recipe-book)
  (:requirements :strips)
  (:predicates
    (in-office ?nutritionist)
    (in-kitchen ?nutritionist)
    (researched-peanut-butter ?nutritionist)
    (has-recipe-draft ?nutritionist ?recipe)
    (has-tested-recipe ?nutritionist ?recipe)
    (computer-charged)
    (has-ingredients ?recipe)
  )

  (:action research_ingredient
    :parameters (?nutritionist ?ingredient)
    :precondition (and 
                    (in-office ?nutritionist))
    :effect (and 
              (researched-peanut-butter ?nutritionist))
  )

  (:action develop_recipe
    :parameters (?nutritionist ?ingredient)
    :precondition (and 
                    (researched-peanut-butter ?nutritionist) 
                    (computer-charged))
    :effect (and 
              (has-recipe-draft ?nutritionist ?ingredient) 
              (not (computer-charged)))
  )

  (:action test_recipe
    :parameters (?nutritionist ?recipe)
    :precondition (and 
                    (in-kitchen ?nutritionist) 
                    (has-recipe-draft ?nutritionist ?recipe) 
                    (has-ingredients ?recipe))
    :effect (and 
              (has-tested-recipe ?nutritionist ?recipe))
  )

  (:action finalize_recipe
    :parameters (?nutritionist ?tested_recipe)
    :precondition (and 
                    (has-tested-recipe ?nutritionist ?tested_recipe) 
                    (computer-charged))
    :effect (and 
              (not (computer-charged)) 
              (not (has-tested-recipe ?nutritionist ?tested_recipe)))
  )
\end{Verbatim}
\end{minipage}
\end{tcolorbox}

\subsection{Trajectory Data}

\begin{tcolorbox}[colback=gray!10!white,colframe=gray!50!white,title=Trajectory Data]
\textbf{\textit{(User)}} You are a nutritionist tasked with creating a new healthy recipe book that incorporates peanut butter powder as a key ingredient. Your environment consists of your office, a computer for research and writing, a kitchen for testing recipes, and a collection of ingredients including peanut butter powder. The computer can be used to write and save recipes, but it needs to be charged after every three hours of use.\\
\\
The actions defined in this domain include:\\
- \*\*research\_ingredient <location> <ingredient> <nutritionist>\*\*: This action allows the nutritionist to research peanut butter powder at the office. The effect of this action is that the nutritionist gains knowledge about the nutritional benefits and potential uses of peanut butter powder in recipes.\\
- \*\*develop\_recipe <location> <computer> <nutritionist> <ingredient>\*\*: This action allows the nutritionist to create a new recipe using peanut butter powder. The precondition for this action is that the nutritionist has researched peanut butter powder. The effect of this action is a draft of a new recipe that incorporates peanut butter powder.\\
- \*\*test\_recipe <location> <kitchen> <nutritionist> <recipe>\*\*: This action allows the nutritionist to test the drafted recipe in the kitchen. The preconditions for this action are that the nutritionist has developed a recipe and has all the necessary ingredients, including peanut butter powder. The effect of this action is feedback on the recipe's taste, nutritional value, and overall feasibility.\\
- \*\*finalize\_recipe <location> <computer> <nutritionist> <tested\_recipe>\*\*: This action allows the nutritionist to finalize the recipe after testing and receiving feedback. The preconditions for this action are that the nutritionist has tested the recipe and the computer is charged. The effect of this action is a completed recipe that is ready to be included in the healthy recipe book.\\
\\
You have the following restrictions on your actions:\\
- You can only research peanut butter powder if you are in the office where your research materials are located.\\
- You can only develop a recipe after researching peanut butter powder and must have a charged computer.\\
- You can only test a recipe in the kitchen if you have developed a recipe and have all the necessary ingredients.\\
- You can only finalize a recipe after testing it in the kitchen and receiving feedback, and if your computer is charged.\\
Goal: The goal is to satisfy the following conditions: The computer is charged. jordan has tested the recipe almond\_butter\_bars.\\
Observation:The computer is charged. The recipe almond\_butter\_bars has all the necessary ingredients. jordan has a recipe draft for almond\_butter\_bars. jordan has researched peanut butter. jordan is in the kitchen.\\
\textbf{\textit{(Assistant)}} Action: jordan tests the recipe almond\_butter\_bars.\\
\textbf{\textit{(User)}} Observation: The computer is charged. The recipe almond\_butter\_bars has all the necessary ingredients. jordan has a recipe draft for almond\_butter\_bars. jordan has researched peanut butter. jordan has tested the recipe almond\_butter\_bars. jordan is in the kitchen.\\
\textbf{\textit{(Assistant)}} Action: jordan develops a recipe using almond\_butter\_bars.\\
\end{tcolorbox}


\end{document}